\pdfoutput=1
\documentclass[11pt]{article}

\usepackage[final]{acl}

\usepackage{times}
\usepackage{latexsym}

\usepackage[T1]{fontenc}
\usepackage[utf8]{inputenc}

\usepackage{microtype}

\usepackage{inconsolata}

\usepackage{graphicx}

\usepackage{times}
\usepackage{latexsym}
\usepackage{enumitem}

\usepackage{listings}
\usepackage{tabularx}
\usepackage{subcaption}
\usepackage{adjustbox}
\usepackage{amsmath}
\usepackage{algorithm}
\usepackage{algpseudocode}
\usepackage[longtable]{multirow}
\usepackage{longtable}
\usepackage{multicol}
\usepackage{caption}
\usepackage{subcaption}
\usepackage{float}
\usepackage{makecell}
\usepackage{amsmath}
\usepackage{amsthm}
\usepackage{amsfonts}
\usepackage{amssymb}
\graphicspath{{./image/}}
\usepackage{booktabs}
\usepackage{kotex} % for korean
\usepackage{tabularray}
\usepackage{url}
\usepackage[normalem]{ulem}
\usepackage{tcolorbox}
\usepackage{pifont}
\usepackage{pdfpages}
\usepackage{color}
\usepackage{tikz}
\usepackage{circuitikz}
\usepackage{hyperref}
\usepackage{arydshln}
\usepackage{array}

\lstdefinestyle{SQL}{
  language=SQL,
  basicstyle=\ttfamily\small,
  breaklines=true,
  keywordstyle=\color{blue},
  stringstyle=\color{red},
  commentstyle=\color{dkgreen},
  morecomment=[l]{\#},
  numbers=none % Remove line numbers
}

\definecolor{Byzantine}{rgb}{0.74, 0.2, 0.64}
\definecolor{Mint}{rgb}{0.24, 0.71, 0.54}
\definecolor{Amaranth}{rgb}{0.9, 0.17, 0.31}
\definecolor{lightcyan}{RGB}{230,255,255}
\definecolor{lightyellow}{RGB}{255,255,224}
\definecolor{lightgreen}{RGB}{240,255,240}
\definecolor{ForestGreen}{RGB}{97,255,77}

\title{FLEX: Expert-level False-Less EXecution Metric \\for Reliable Text-to-SQL Benchmark}

\author{
 \textbf{Heegyu Kim\textsuperscript{1}},
 \textbf{Taeyang Jeon\textsuperscript{1}},
 \textbf{Seunghwan Choi\textsuperscript{1}},
 \textbf{Seungtaek Choi\textsuperscript{3}},
 \textbf{Hyunsouk Cho\textsuperscript{1,2}\thanks{~ Corresponding author}}
\\
 \textsuperscript{1}Department of Artificial Intelligence,\\
 \textsuperscript{2}Department of Software and Computer Engineering,\\
 Ajou University, Suwon 16499, Republic of Korea\\
 \textsuperscript{3}Yanolja, Republic of Korea
\\
}

\begin{document}
\maketitle

\newcommand{\se}{{\it SE}}%
\newcommand{\eg}{{\it e.g.}}%
\newcommand{\ie}{{\it i.e.}}%
\newcommand{\etal}{{\it et al.}}%
\newcommand{\etc}{{\it etc}}%

\newcommand{\argmin}{\operatornamewithlimits{argmin}}
\newcommand{\argmax}{\operatornamewithlimits{argmax}}

\newcommand{\com}{\textcolor{red}}

\def\geotextual{{spatial-keyword}}
\def\geospatial{geo-spatial}
\def\PI{\mathcal{P}}
\newcommand{\XXP}[1]{{\PI(#1)}}
\def\XXQEO{\emph{$Q_1$}}
\def\kNN{\textsc{$k$NN}}
\def\XXD{\mathcal{D}}
\def\XXT{\mathcal{T}}
\newcommand{\XXDN}[0]{{D}}
\newcommand{\XXTN}[0]{{T}}
\def\Base{\textsc{Base}}
\def\TopK{\textsc{Top-$k$}}
\def\tag{{keyword}}
\def\Query{{Query}}
\newcommand{\ttag}[1]{{`#1'}}

\newcommand{\vanilla}{\textbf{Vanilla}}
\newcommand{\simple}{\textbf{FP-Simple}}

% Caligraphy style
\newcommand{\mcal}[1]{{\cal{#1}}}
\newcommand{\calA}{\mbox{${\cal A}$}}
\newcommand{\calB}{\mbox{${\cal B}$}}
\newcommand{\calC}{\mbox{${\cal C}$}}
\newcommand{\calD}{\mbox{${\cal D}$}}
\newcommand{\calE}{\mbox{${\cal E}$}}
\newcommand{\calF}{\mbox{${\cal F}$}}
\newcommand{\calG}{\mbox{${\cal G}$}}
\newcommand{\calH}{\mbox{${\cal H}$}}
\newcommand{\calI}{\mbox{${\cal I}$}}
\newcommand{\calJ}{\mbox{${\cal J}$}}
\newcommand{\calK}{\mbox{${\cal K}$}}
\newcommand{\calL}{\mbox{${\cal L}$}}
\newcommand{\calM}{\mbox{${\cal M}$}}
\newcommand{\calN}{\mbox{${\cal N}$}}
\newcommand{\calO}{\mbox{${\cal O}$}}
\newcommand{\calP}{\mbox{${\cal P}$}}
\newcommand{\calQ}{\mbox{${\cal Q}$}}
\newcommand{\calR}{\mbox{${\cal R}$}}
\newcommand{\calS}{\mbox{${\cal S}$}}
\newcommand{\calT}{\mbox{${\cal T}$}}
\newcommand{\calU}{\mbox{${\cal U}$}}
\newcommand{\calV}{\mbox{${\cal V}$}}
\newcommand{\calW}{\mbox{${\cal W}$}}
\newcommand{\calX}{\mbox{${\cal X}$}}
\newcommand{\calY}{\mbox{${\cal Y}$}}
\newcommand{\calZ}{\mbox{${\cal Z}$}}

%experiments
\newcommand{\LM}{\calL\calM}
\newcommand{\SR}{{Self-Refine}}
\newcommand{\ours}{\textbf{FLEX}}
\newcommand{\llmsqlsolver}{\textit{LLM-SQL-Solver}}
\newcommand{\prometheus}{\textit{Prometheus}}
\newcommand{\prometheustwo}{\textit{Prometheus-2}}
\newcommand{\miniatureandmull}{\textit{Miniature-and-Mull}}
\newcommand{\explainandcompare}{\textit{Explain-and-Compare}}

\newcommand{\db}{$D$}
\newcommand{\schema}{$S$}
\newcommand{\nlqset}{$X$}
\newcommand{\nlq}{$x$}
\newcommand{\sqlgt}{$Q_{gt}(x)$}
\newcommand{\sqlgen}{$Q_{gen}(x)$}
\newcommand{\exgt}{$R_{gt}(x)$}
\newcommand{\exgen}{$R_{gen}(x)$}

\newcommand{\todoc}[2]{{\textcolor{#1}{\textbf{#2}}}}
\newcommand{\todoorange}[1]{\todoc{orange}{\textbf{[[#1]]}}}
\newcommand{\hist}[1]{\todoorange{hist: #1}}

\newtcolorbox{defin}{enhanced,
	attach boxed title to top left={xshift=-4mm},boxrule=0pt,after skip=1cm,before skip=1cm,right skip=0cm,breakable,fonttitle=\bfseries,toprule=0pt,bottomrule=0pt,rightrule=0pt,leftrule=4pt,arc=0mm,skin=enhancedlast jigsaw,sharp corners,colframe=gree,colbacktitle=gre,boxed title style={
		frame code={ 
			\fill[gre](frame.south west)--(frame.north west)--(frame.north east)--([xshift=3mm]frame.east)--(frame.south east)--cycle;
			\draw[line width=1mm,gre]([xshift=2mm]frame.north east)--([xshift=5mm]frame.east)--([xshift=2mm]frame.south east);
			
			\draw[line width=1mm,gre]([xshift=5mm]frame.north east)--([xshift=8mm]frame.east)--([xshift=5mm]frame.south east);
			\fill[green!40](frame.south west)--+(4mm,-2mm)--+(4mm,2mm)--cycle;
		}
	}
}

\begin{abstract}
%short version
Text-to-SQL systems have become crucial for translating natural language into SQL queries in various industries, enabling non-technical users to perform complex data operations. The need for accurate evaluation methods has increased as these systems have grown more sophisticated. However, the Execution Accuracy (EX), the most prevalent evaluation metric, still shows many false positives and negatives. Thus, this paper introduces \ours\ \textbf{(False-Less EXecution)}, a novel approach to evaluating text-to-SQL systems using large language models (LLMs) to emulate human expert-level evaluation of SQL queries. Our metric improves agreement with human experts (from 62 to 87.04 in Cohen's kappa) with comprehensive context and sophisticated criteria. Our extensive experiments yield several key insights: (1) Models' performance increases by over 2.6 points on average, substantially affecting rankings on Spider and BIRD benchmarks; (2) The underestimation of models in EX primarily stems from annotation quality issues; and (3) Model performance on particularly challenging questions tends to be overestimated. 
% Our re-evaluation of all publicly available models provides several insights. The models' performance increases by over 2.6 points on average, significantly affecting model rankings on Spider and BIRD benchmarks. We found that the underestimation in EX mainly results from annotation quality issues, while performance on particularly challenging questions tends to be overestimated. 
This work contributes to a more accurate and nuanced evaluation of text-to-SQL systems, potentially reshaping our understanding of state-of-the-art performance in this field.
\end{abstract}
\section{Introduction}
Text-to-SQL systems, which translate natural language questions into SQL queries, have become vital across various industries by democratizing data access and facilitating data-driven decision-making~\cite{shi2024survey,hong2024nextgenerationdatabaseinterfacessurvey}. As these systems grow more complex, accurate and efficient evaluation methods are increasingly important.
While initial evaluations relied on human experts, this approach proved too time-consuming and costly for large-scale assessments. To address this benchmarks like Spider~\cite{spider} introduced the Execution Accuracy (EX) metric, which has been widely adopted by others such as BIRD~\cite{BIRD}.

\begin{table}[!t]
\centering
\caption{BIRD top 10 leaderboard re-ranked by \ours. $\Delta$ denotes underestimation error (FLEX $-$ EX).}
\label{tab:tab_intro_bird}
\adjustbox{max width=\columnwidth}{
\begin{tabular}{llccc}
\toprule
\textbf{Rank} & \textbf{Model} & \textbf{FLEX} & \textbf{EX} & \textbf{$\Delta$} \\
\midrule
\phantom{0}1\phantom{0}(\textcolor{red}{↑2}) & SuperSQL {\includegraphics[height=0.4cm]{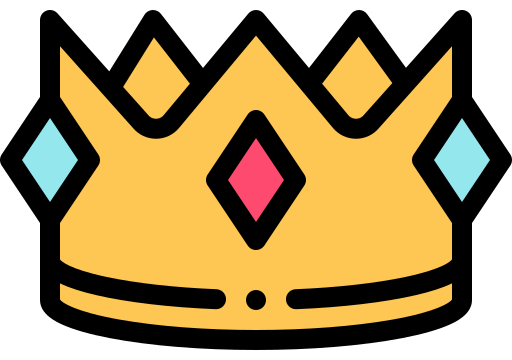}} & \textbf{64.08} & 57.37 & $+6.71$ \\
\phantom{0}2\phantom{0}(\textcolor{blue}{↓1}) & CHESS-GPT-4o-mini & 62.71 & \textbf{59.13} & $+3.59$ \\
\phantom{0}3\phantom{0}(\textcolor{red}{↑2}) & TA-ACL & 59.97 & 55.67 & $+4.30$ \\
\phantom{0}4\phantom{0}(\textcolor{red}{↑3}) & DAIL\_SQL\_9-SHOT\_MP & 59.26 & 53.52 & $+5.74$ \\
\phantom{0}5\phantom{0}(\textcolor{red}{↑4}) & DAIL\_SQL\_9-SHOT\_QM & 58.47 & 53.06 & $+5.41$ \\
\phantom{0}5\phantom{0}(\textcolor{blue}{↓3}) & DTS-SQL-BIRD-GPT4o & 58.47 & 58.08 & $+0.39$ \\
\phantom{0}7\phantom{0}(\textcolor{blue}{↓3}) & SFT\_CodeS\_15B\_EK & 56.98 & 56.52 & $+0.46$ \\
\phantom{0}8\phantom{0}(\textcolor{blue}{↓2}) & SFT\_CodeS\_7B\_EK & 53.59 & 54.89 & $-1.30$ \\
\phantom{0}9\phantom{0}(\textcolor{blue}{↓1}) & SFT\_CodeS\_3B\_EK & 53.26 & 53.46 & $-0.20$ \\
10\phantom{0}(\textcolor{red}{↑2}) & DAIL\_SQL & 51.83 & 45.89 & $+5.93$ \\
\bottomrule
\end{tabular}
}
\end{table}

However, the EX metric, which evaluates queries based on their execution result, has significant limitations that can lead to assessments diverging from human experts. Our analysis reveals that EX incorrectly validates queries due to coincidental database states or unfairly penalizes correct ones because of ambiguous questions, thus misestimating a model's ability to generate correct and valid SQL queries~\cite{pourreza2023evaluating}.

Although LLM-based evaluation methods~\cite{kim2023prometheus,zheng2023judgingllmasajudgemtbenchchatbot,zhao2024llmsqlsolverllmsdeterminesql} have been proposed, they exhibit suboptimal performance due to limited context.
They fail to utilize crucial contextual information, struggle with noisy ground truth issues, and employ ambiguous criteria inappropriate for text-to-SQL evaluation. 
Consequently, they even underperform traditional EX metric when evaluated against established text-to-SQL benchmarks.

To address these issues, we propose \ours~\textbf{(False-Less EXecution)}, a novel approach leveraging LLMs to emulate expert-level evaluation of SQL queries. \ours~is specifically designed for evaluating text-to-SQL tasks and overcomes existing limitations through comprehensive context analysis, considering the natural language question, database schema, and external knowledge. It also exploits elaborately devised evaluation criteria with detailed guidelines for assessing query correctness. Our approach is validated through strong consistency between \ours~judgments and human expert evaluations, demonstrating significantly higher agreement (Cohen's kappa of 87.04) than the existing EX metric (62.00) and outperforming previous LLM-based methods.

% To address these issues, we propose \ours~\textbf{(False-Less EXecution)}, a novel approach leveraging LLMs to emulate expert-level evaluation of SQL queries. \ours~is specifically designed for text-to-SQL tasks and overcomes existing limitations through innovations like comprehensive context analysis that considers the natural language question, database schema, and external knowledge. 

% We developed sophisticated evaluation criteria with detailed guidelines for assessing query correctness, addressing common pitfalls in text-to-SQL tasks. \ours~offers robust handling of noisy ground truth, correctly evaluating queries even when the provided ground truth is inaccurate or ambiguous. Our approach is validated through strong consistency between \ours~judgments and human expert evaluations, demonstrating significantly higher agreement (Cohen's kappa of 87.04) than the existing EX metric (62.00) and outperforming previous LLM-based methods.

\begin{figure}[!t]
\includegraphics[width=\linewidth]{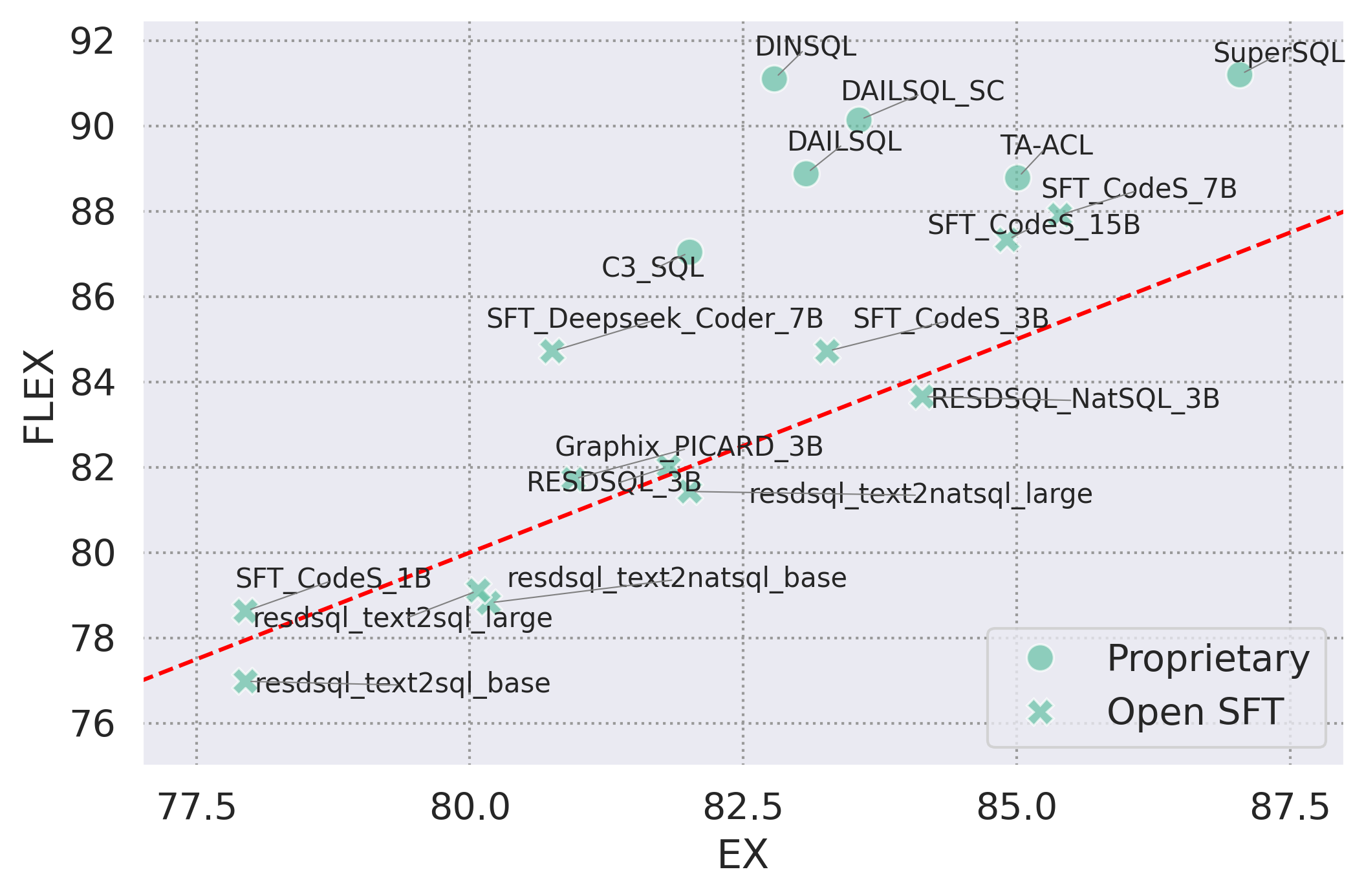}
\vspace{-20pt}
\caption{Performance comparison of EX vs. FLEX metrics on Spider benchmark. The red identity line shows an equivalent score.}
\label{fig:scatter-ex-flex-spider-cropped}
\end{figure}
To demonstrate the practical impact of \ours, we re-evaluated 50 publicly available text-to-SQL models on the Spider and BIRD\footnote{Our study utilized the 20240627 version of the BIRD-dev set, yielding different EX scores from prior studies due to modified ground truth.} benchmarks, revealing several key insights. 
% As shown in Table~\ref{tab:tab_intro_bird}, our analysis indicates significant shifts in model rankings, particularly for BIRD, providing a more accurate assessment of model capabilities. 
As shown in Table~\ref{tab:tab_intro_bird}, significant shifts in model rankings, particularly for BIRD, provide a more accurate assessment of model capabilities.
Additionally, \ours~categorized the error cases and identified noisy annotations as a primary factor in underestimating model performance in current benchmarks. As illustrated in Fig.~\ref{fig:scatter-ex-flex-spider-cropped}, \ours~scores are generally higher than EX scores, suggesting that \ours~captures previously underestimated aspects of model capability. Furthermore, \ours~detected models are overestimated in BIRD's challenging questions, highlighting areas for future research focus.

Our key contributions are as follows: 1) we identify key limitations in current text-to-SQL evaluation metrics, especially the Execution Accuracy (EX) metric, and 2) we introduce \ours, a novel LLM-based evaluation method that aligns with human expert reasoning and shows better agreement with expert assessments. 3) By re-evaluating 50 text-to-SQL models using \ours, we uncover significant ranking changes and provide a clearer understanding of benchmarks and state-of-the-art models, emphasizing the need for advanced evaluation methods. \ours~framework is publicly available~\footnote{\url{https://github.com/HeegyuKim/FLEX}}.

\section{Related Works}
% \subsection{Conventional Metrics}
\textbf{Exact Matching~(EM) and Execution Accuracy~(EX)} were proposed to evaluate text-to-SQL systems in the Spider~\cite{spider}. 
EM assesses the syntax-level equivalence of two queries; however, it is prone to high false negative rates because the same logical intent can be expressed in various query formulations.
Therefore, EX has been utilized as a primary metric for query accuracy in BIRD~\cite{BIRD}, focusing on the execution results rather than the syntactic form.
Despite this improvement, EX still suffers from false negatives due to annotation quality issues and may produce false positives by rewarding incorrect queries that coincidentally yield the same results as the ground truth. These limitations underscore the necessity for more robust semantic evaluation approaches.

\textbf{LLM-based evaluation} has gained popularity as an alternative to human evaluators for assessing human preference in natural language generation models. Frameworks like MT-Bench~\cite{zheng2023judgingllmasajudgemtbenchchatbot} and AlpacaEval~\cite{alpaca_eval} utilize LLMs as judges to provide faster and more cost-effective evaluations than human assessments. However, these frameworks primarily rely on the proprietary GPT-4~\cite{openai2024gpt4technicalreport}, which poses reproducibility risks since providers may alter or discontinue models without notice. In response, \citeauthor{kim2023prometheus} introduced the open-source LLM judge \prometheus, offering a reproducible and cost-free alternative at GPT-4's performance level. Nonetheless, \prometheus\ is mainly focused on general human preference evaluations rather than specialized tasks like text-to-SQL evaluation.
To specifically leverage LLMs for text-to-SQL evaluation, \citeauthor{zhao2024llmsqlsolverllmsdeterminesql} proposed \llmsqlsolver. Without distinguishing between the false positives and false negatives, this approach showcases the capability of LLMs to assess execution equivalence between two queries using two prompting strategies: \miniatureandmull, evaluating semantic equivalence between two queries, and \explainandcompare, asking LLMs to compare their significant logical differences. Despite its advancements, \llmsqlsolver\ remains insufficient to fully replace EX as an evaluation metric. We will investigate these detailed limitations in section~\ref{sec:llm_replace_ex}.
\section{Preliminaries}
This section introduces the notation and definitions used throughout the paper. We begin by establishing the following conventions: 
Let $D$ denote a database, $S$ its schema, and $X$ a set of natural language questions. For each question $x \in X$, we define contextual information $\mathbb{C}$ to assist in the evaluation of metrics, including $Q_{gt}(x)$ as the ground truth SQL query for question $x$, and $Q_{gen}(x)$ as the SQL query generated by a text-to-SQL model for $x$. 
We then execute these queries on the database $D$ to obtain their execution result sets. Specifically, we define $R_{gt}(x) = \text{Execute}(Q_{gt}(x), D)$ as the result of the ground truth query and $R_{gen}(x) = \text{Execute}(Q_{gen}(x), D)$ as the result of the generated query.
% We then execute these queries on the database $D$ to obtain result sets where $R_{gen}(x)$ is the result set of the generated query~(\ie, $R_{gen}(x) = \text{Execute}(Q_{gen}(x), D)$) and $R_{gt}(x)$ is the result set of the ground truth query~(\ie, $R_{gt}(x) = \text{Execute}(Q_{gt}(x), D)$).
% \begin{align}
% R_{gen}(x) &= Execute(Q_{gen}(x), D) \\
% R_{gt}(x) &= Execute(Q_{gt}(x), D)
% \end{align}
We define evaluation metrics: EM, EX, and LLM-based evaluation~(LX) as: 
\begin{align}
& \text{EM} = \frac{1}{|X|}\sum_{x \in X}\mathbb{I}\left(Q_{gen}(x) = Q_{gt}(x) \right) \\
& \text{EX} = \frac{1}{|X|}\sum_{x \in X}\mathbb{I}\left( R_{gen}(x) = R_{gt}(x) \right) \\
% & LX = \frac{1}{|X|}\sum_{x \in X}\mathbb{I}\left( LLM(x, \mathbb{C}) = Correct\right) \\
% & LLM(x, \mathbb{C}) = \begin{cases}
% Correct & \text{if } x \leftrightarrow_{\mathbb{C}} Q_{gen}(x) \\
% Incorrect  & \text{if } x \nleftrightarrow_{\mathbb{C}} Q_{gen}(x)
% \end{cases}
% & \text{LX} = \frac{1}{|X|}\sum_{x \in X}\mathbb{I}\left( x \leftrightarrow_{\mathbb{C}} Q_{gen}(x) \right)
& \text{LX} = \frac{1}{|X|}\sum_{x \in X}\mathbb{I}\left( x \underset{\mathbb{C}}{\leftrightarrow} Q_{gen}(x) \right)
\end{align}
where $\mathbb{I}(\cdot)$ is the indicator function that returns 1 if the condition inside the parentheses is true, and 0 otherwise.
The notation $x \underset{\mathbb{C}}{\leftrightarrow} Q_{gen}(x)$ denotes that the generated query $Q_{gen}(x)$ semantically matches the question $x$ under the contextual information $\mathbb{C}$ based on an LLM.

For EM, the condition checks if $Q_{gen}(x)$ exactly matches the $Q_{gt}(x)$ syntactically.
For EX, the condition checks if the $R_{gen}(x)$ exactly matches the $R_{gt}(x)$.
However, LX handles semantic match between \sqlgen~and question \nlq~under contextual information. $\mathbb{C}$ may vary depending on the evaluation method.
These metrics produce a score between 0 and 1, where 1 indicates perfect performance (100\% accuracy) and 0 indicates complete failure (0\% accuracy).

\section{Analyzing the Limitations of Current Text-to-SQL Evaluation Methods}
\label{sec:motivation}
In this section, we address key research questions to investigate the shortcomings of existing text-to-SQL evaluation metrics. We aim to understand the types of errors these metrics produce and explore whether LLMs can offer a better alternative.

\subsection{What Types of Errors Occur in Text-to-SQL Evaluation?}
\label{sec:pre_error_analysis}
To identify the limitations of current evaluation methods, we first examine the errors that arise when using the Execution Accuracy (EX) metric: false positives (FP) and false negatives (FN) using Table~\ref{tab:tab_fp_fn_example}.
% In this subsection, we briefly explain the evaluation process of EX and investigate its error cases. \hist{EX의 evaluation process 를 설명한다기엔 다음 한줄이 전부라 조금 아쉽}
% Text-to-SQL benchmarks provide natural language questions with database schema and external knowledge to generate queries from models. 
% For assessment, benchmarks primarily employ EX, which executes the generated query and compares the result with the ground truth.
% However, EX suffers from two significant limitations: false positives (FP) and false negatives (FN).

\begin{table}[t]
\caption{Illustrated examples of false positive / negative.}

\resizebox{\columnwidth}{!}{%\

% \renewcommand{\arraystretch}{0.9}

% \begin{tabular}{ll}

% \hline

% \textcolor{orange}{\nlq}           & \textit{Who has the highest score?} \\ \hdashline

% \textcolor{blue}{\sqlgt}         & SELECT fname, lname FROM student \\

%                                        & ORDER BY score DESC LIMIT 1 \\

%                                        & > Emily, Carter \\ \hdashline

% EX 1:                              & SELECT fname, lname FROM student \\ 

% \textcolor{Byzantine}{FP}  & \textbf{\textcolor{Byzantine}{WHERE age < 19}} \\ 

%                                        & ORDER BY score DESC LIMIT 1 \\

%                                        & > Emily, Carter \\ 

%                                        & \textcolor{teal}{\# WHERE clause is unnecessary.} \\ \hdashline

% \eg,~2                              & SELECT \textbf{\textcolor{Amaranth}{lname, fname}} FROM student \\

% \textcolor{Amaranth}{FN 1} & ORDER BY score DESC LIMIT 1 \\ 

%                                        & > Carter, Emily \\ 

%                                        & \textcolor{teal}{\# The column order is different} \\ \hdashline

% \eg,~3                              & SELECT fname, lname FROM student \\

% \textcolor{Amaranth}{FN 2} & WHERE score == (SELECT      \\

%                                        & ~ MAX (score) FROM student)                        \\

%                                        & > Carter, Emily | Liam, Thompson                          \\

%                                        & \textcolor{teal}{\# Two students have the highest score.} \\ \hline

% \end{tabular}
\begin{tabular}{l} 
\hline
\textit{\textbf{Question: Who has the highest score?}}                                                                                                                                                                              \\ 
\hline
\textbf{Ground Truth Query}                                                                                                                                                                                                         \\ 
\hdashline[1pt/1pt]
SELECT fname, lname FROM student                                                                                                                                                                                                    \\
ORDER BY score DESC LIMIT 1                                                                                                                                                                                                         \\
\textgreater{} Emily, Carter                                                                                                                                                                                                                      \\ 
\hline
\textbf{Example 1: \textcolor{Byzantine}{False Positive}}                                                                                                                                                                                                  \\ 
\hdashline[1pt/1pt]
SELECT fname, lname FROM student~                                                                                                                                                                                                   \\
\textcolor{Byzantine}{\textbf{WHERE age \textless{} 19~}}\textcolor[rgb]{0,0.502,0.502}{\# Unnecessary condition.}  \\
ORDER BY score DESC LIMIT 1                                                                                                                                                                                                         \\
\textgreater{} Emily, Carter                                                                                                                                                                                                                      \\ 
\hline
\textbf{Example 2: \textcolor{Amaranth}{False Negative}}                                                                                                                                                                                                  \\ 
\hdashline[1pt/1pt]
SELECT \textcolor{Amaranth}{\textbf{lname, fname}} FROM student                                                                                                                                                                                           \\
ORDER BY score DESC LIMIT 1                                                                                                                                                                                                         \\
\textgreater{} Carter, Emily~\textcolor[rgb]{0,0.502,0.502}{\# The column order is different}                                                                                                                                       \\ 
\hline
\textbf{Example 3: \textcolor{Amaranth}{False Negative}}                                                                                                                                                                                                  \\ 
\hdashline[1pt/1pt]
SELECT fname, lname FROM student                                                                                                                                                                                                    \\
WHERE score == (SELECT~ MAX (score) FROM student)                                                                                                                                                                                   \\
\textgreater{} Emily, Carter \textbar{} Liam, Thompson~                                                                                                                                                                                           \\
\textcolor[rgb]{0,0.502,0.502}{\# Two students have the highest score.}                                                                                                                                                             \\
\hline
\end{tabular}
}

\label{tab:tab_fp_fn_example}

\end{table}

% \vspace{-20pt} % table과 text 간격 줄이기
% \textbf{False Positives:} Queries with semantically different structures or logic may produce the same execution results due to the current database states, leading to overestimating system performance.
% As described in Table~\ref{tab:tab_fp_fn_example}, Example 1 has an unnecessary condition $age < 19$ but coincidentally produces the same results as the ground truth because the database has no students older than 18. 
% Unlike humans, EX does not assess whether the generated SQL accurately represents the intent of the natural language question, potentially rewarding queries that produce correct results through incorrect means.
% \begin{equation}
% x \nleftrightarrow Q_{gen}(x), \text{but } R_{gen}(x) = R_{gt}(x)
% \end{equation}
% \textbf{False Positives = $x \nleftrightarrow Q_{gen}(x), \text{but } R_{gen}(x) = R_{gt}(x)$.} 
% \noindent\textbf{False Positives} ($x \nleftrightarrow_{\mathbb{C}} Q_{gen}(x), \text{but } R_{gen}(x) = R_{gt}(x)$): 
\noindent\textbf{False Positives} ($x \underset{\mathbb{C}}{\nleftrightarrow} Q_{gen}(x), \text{but } R_{gen}(x) = R_{gt}(x)$):
The generated query \sqlgen~with semantically different structures or logic may produce the same execution results due to the current database states, leading to overestimating system performance. \emph{Example 1} includes an unnecessary condition \texttt{age < 19} but coincidentally produces the same results as the ground truth because the database has no students older than 18. Unlike humans, EX does not assess whether the generated SQL accurately represents the intent of the natural language question, potentially rewarding queries that yield correct results through incorrect means.

% \textbf{False Negatives:} Semantically correct queries that translate natural language questions accurately produce different results from ground truth queries, underestimating system capabilities. 
% Prior studies~\cite{pourreza2023evaluating,wretblad2024understanding} issued annotation quality, potentially yielding false negatives while evaluating.
% They claimed that the Spider and BIRD benchmarks have many annotation issues, at least 15\% per database, including ambiguous questions and incorrectly annotated ground truths.
% Ambiguous questions can be interpreted as multiple queries. 
% For instance, the question in Table~\ref{tab:tab_fp_fn_example} does not specify the output column and asks `Who.'
% Thus, though the Example 2 query has the correct filtering condition, it is evaluated as incorrect because it selects two columns, id and name.
% Also, Table~\ref{tab:tab_fp_fn_example} shows noisy ground truth, which can be issued when multiple students have the same highest score. Example 3 query correctly handles this issue, but it is evaluated as incorrect because its execution result differs from the ground truth.
% \begin{equation}
% x \leftrightarrow Q_{gen}(x), \text{but } R_{gen}(x) \neq R_{gt}(x)
% \end{equation}
% \textbf{False Negatives: $x \leftrightarrow Q_{gen}(x), \text{but } R_{gen}(x) \neq R_{gt}(x)$.} 
% \noindent\textbf{False Negatives} ($x \leftrightarrow_{\mathbb{C}} Q_{gen}(x), \text{but } R_{gen}(x) \neq R_{gt}(x)$): 
\noindent\textbf{False Negatives} ($x \underset{\mathbb{C}}{\leftrightarrow} Q_{gen}(x), \text{but } R_{gen}(x) \neq R_{gt}(x)$): 
Semantically correct queries that accurately translate the natural language questions may produce different results from ground truth queries, underestimating system capabilities. Prior studies~\cite{pourreza2023evaluating,wretblad2024understanding} have raised concerns about annotation quality, potentially leading to false negatives during evaluation. They reported that the Spider and BIRD benchmarks have many annotation issues—at most 49\% at specific databases—including ambiguous questions and incorrectly annotated ground truths.

Ambiguous questions can be translated into multiple queries due to limited instruction about the output structure.
For instance, \emph{Example 2} is marked incorrect simply because \texttt{lname} precedes \texttt{fname} in the column order. This structural rigidity in evaluation can lead to false negatives, potentially overlooking semantically correct queries.
Additionally, \emph{Example 3} shows noisy ground truth, which can occur when multiple students have the same highest score. The query handles this issue correctly but is evaluated as incorrect because its execution result differs from the ground truth.
% These limitations reduce reliability and hinder the development of more robust text-to-SQL systems, potentially leading to deployed systems that fail in real-world scenarios.
% To address these issues, text-to-SQL systems must employ human experts or expert-level LLMs for accurate evaluation.

These limitations reduce reliability and hinder the development of more robust text-to-SQL systems, potentially leading to deployed systems that fail in real-world scenarios. To address these issues, we must explore evaluation methods that align more closely with human judgment.

\subsection{How Closely Does EX Align with Human Expert Evaluations?}
\label{sec:pre_human_study}
% Building on the identified errors, we pose the research question: \textbf{How close is EX to human experts in evaluating SQL query correctness?} To assess this, we conducted a human evaluation comparing the agreement between human experts and EX.
Building on our analysis of identified errors, we conducted a human evaluation study to assess the alignment between EX and human experts in evaluating SQL query correctness.
We randomly sampled 200 query pairs from the BIRD dataset, generated by TA-ACL~\cite{qu2024generationalignitnovel} and SuperSQL~\cite{li2024dawn}. The samples were evenly divided between queries that produced the same result as the ground truth (\emph{Equivalent set}) and those with different results (\emph{Not equivalent set}). Three SQL experts with over three years of experience independently evaluated these queries for semantic correctness. Discrepancies were resolved through consensus. Before reaching a consensus, the inter-annotator agreement measured by Fleiss' kappa was 79.32, indicating strong agreement.

% \textbf{Dataset and Human Annotation:} We randomly sampled 200 query pairs from the BIRD dataset, generated by TA-ACL~\cite{qu2024generationalignitnovel} and SuperSQL~\cite{li2024dawn}. The samples were evenly divided between queries that produced the same result as the ground truth (\emph{Equivalent set}) and those with different results (\emph{Not equivalent set}). Three SQL experts with over three years of experience independently evaluated these queries for semantic correctness. Discrepancies were resolved through consensus. Before reaching a consensus, the inter-annotator agreement measured by Fleiss' kappa was 86.7, indicating almost perfect agreement.
% \textbf{Metrics and Results:} 
We employed Cohen's kappa and accuracy score to measure the agreement between human experts and EX. The agreement between human consensus and EX yielded a Cohen's kappa of 62.0, indicating substantial agreement but leaving much room for improvement, where human annotators identified 21 false positives in the \emph{Not equivalent set} and 17 false negatives in the \emph{Equivalent set}.

These findings raise concerns about the reliability of the EX metric and underscore the need for evaluation methods that better reflect human judgment by considering semantic correctness and providing a more expert-level assessment of query accuracy. Appendix~\ref{sec:appendix_errors} provides real examples illustrating these discrepancies.
\begin{table}
\centering
\caption{Agreement between human consensus and other evaluation methods. \textbf{Acc} denotes overall accuracy score, \textbf{EQ} denotes \emph{Equivalent set} accuracy score, \textbf{NEQ} denotes \emph{Not equivalent set} accuracy score.}
\label{tab:tab_pre_human_agreement}
\adjustbox{max width=1\linewidth}{
\begin{tblr}{
  hline{1,6} = {-}{0.08em},
  column{even} = {c},
  column{3} = {c},
  column{5} = {c},
  vline{2} = {-}{},
  hline{1-2} = {-}{},
  hline{3} = {-}{dashed},
  rowsep = 0.5pt,
}
\textbf{Model}    & \textbf{Kappa} & \textbf{Acc}  & \textbf{EQ} & \textbf{NEQ} \\
EX                & \textbf{62.00} & \textbf{81.0} & \textbf{79} & \textbf{83}  \\
LLM-SQL-Solver    & 52.29          & 76.5          & 70          & \textbf{83}  \\
Prometheus-2-7B   & \uline{61.14}  & \uline{80.5}  & \uline{78}  & \textbf{83}  \\   
Prometheus-2-8x7B & 60.66          & 80.0          & 78          & \uline{82}     
\end{tblr}
}
\end{table}

% \begin{table}
% \centering
% \caption{Agreement between human consensus and other evaluation methods. \textbf{Acc} denotes overall accuracy score, \textbf{EQ} denotes \emph{equivalent set} accuracy score, \textbf{NEQ} denotes \emph{not equivalent set} accuracy score.}
% \label{tab:tab_pre_human_agreement}
% \begin{tblr}{
%   column{even} = {c},
%   column{3} = {c},
%   column{5} = {c},
%   vline{2} = {-}{},
%   hline{1,10} = {-}{0.08em},
%   hline{2} = {-}{},
%   hline{3} = {-}{dashed},
% }
% \textbf{Model}           & \textbf{Kappa} & \textbf{Acc}  & \textbf{EQ} & \textbf{NEQ} \\
% EX                       & 62.00          & 81.0          & 79          & 83           \\
% Prometheus w/ GPT-4o-A~  & \uline{67.85}  & \uline{84.0}  & \textbf{83} & \uline{85}   \\
% Prometheus w/ GPT-4o-C   & \textbf{70.97} & \textbf{85.5} & \uline{82}  & \textbf{89}  \\
% LLM-SQL-Solver w/ GPT-4o & 52.29          & 76.5          & 70          & 83           \\
% Prometheus-2-7B-A        & 52.91          & 76.5          & 71          & 82           \\
% Prometheus-2-7B-C        & 61.14          & 80.5          & 78          & 83           \\
% Prometheus-2-8x7B-A      & 40.47          & 70.5          & 67          & 74           \\
% Prometheus-2-8x7B-C      & 60.66          & 80.0          & 78          & 82           
% \end{tblr}
% \end{table}

\subsection{Can LLMs Replace EX in Evaluating Text-to-SQL Systems?}
\label{sec:llm_replace_ex}
% Given the limitations of EX and its misalignment with human judgment, we explore the research question: \textbf{Can LLMs replace EX for evaluating text-to-SQL systems?}
Given EX's limitations and misalignment with human judgment, we explore LLMs as a potential alternative for evaluating text-to-SQL systems.
We thus employed two LLM-based evaluation methods: 1) \prometheustwo~\cite{kim2024prometheus}, an open LM judge, and 2) \llmsqlsolver~\cite{zhao2024llmsqlsolverllmsdeterminesql} with GPT-4o~\cite{openai2024gpt4technicalreport}, a prompting method utilizing a proprietary LLM. 
\textbf{Human Experts vs. LLMs:} Table~\ref{tab:tab_pre_human_agreement} reveals that existing LLM-based evaluation methods did not surpass EX performance. \prometheustwo\ showed a marginally lower agreement than EX, with a Cohen's kappa of 61.14, while \llmsqlsolver\ performed considerably worse, even when using a flagship proprietary LLM. We propose three main factors contributing to the underperformance of current LLM-based evaluation methods:

\begin{itemize}[noitemsep,topsep=5pt,parsep=2pt,partopsep=0pt,leftmargin=*]
    \item \textbf{Limited Context and Noisy Ground Truth:} \llmsqlsolver~neglects contextual information $\mathbb{C}$, including the natural language question, external knowledge, and execution results. It solely compares queries \sqlgen~to \sqlgt~under schema \schema, disregarding whether \sqlgen~accurately represents question $x$, thus limiting its ability to detect false positives and negatives. This issue is particularly critical in the Spider and BIRD benchmarks due to annotation quality concerns discussed in Section~\ref{sec:pre_error_analysis}.
    
    \item \textbf{Ambiguous Criteria:} The \llmsqlsolver~may underperform even with advanced LLMs due to unspecified evaluation criteria. Its \explainandcompare~strategy uses vague prompts like "Are there significant differences?" Similarly, \prometheustwo~employs a 1-5 rubric scoring system, but designing optimal rubrics becomes heuristic and ambiguous, especially for scores between 2 and 4 in binary decisions (correct and incorrect)\footnote{Therefore, we empirically selected rubrics with a threshold ($\geq4$) yielding the highest agreement scores.} We claim that LLM should evaluate SQL queries using deeper reasoning steps and several binary criteria; otherwise, the LLM shows low evaluation performance from the ambiguous criteria.
    % We propose an optimal prompt incorporating specific evaluation criteria for false positives and negatives to enhance LLM-based evaluators' performance.
\end{itemize}

Our study empirically demonstrates that prior LLM-based evaluation methods are insufficient compared to EX for evaluating text-to-SQL systems. This highlights the need for a more advanced evaluation paradigm that leverages LLMs effectively while addressing their current limitations.
% \textbf{1) Limited Context:} \llmsqlsolver\ does not utilize additional context, including the natural language question, external knowledge, and execution results, and just compares the generated query to the ground truth query.
% \textbf{2) Noisy Ground Truth:} Due to the previous reason, \llmsqlsolver\ ignores whether the generated query represents the question correctly, limited to capturing the false positives. It is crucial in Spider and BIRD benchmarks due to the annotation quality concerns addressed in section~\ref{sec:pre_error_analysis}.
% \textbf{3) Ambiguous Criteria}: The \llmsqlsolver\ is potentially underperformed even in the flagship LLM due to unspecified criteria. 
% \explainandcompare\ strategy does not specify the evaluation criteria and prompts LLM to ask, ``Are there significant differences?''
% On the basis, as described in table~\ref{tab:tab_pre_human_agreement}, \prometheustwo\ with customized template~(C) benefits the Cohen's kappa score, +8.23, than ambiguous criteria~(A). This implies there is room for improvement in cutting-edge LLMs through progressive methodology.
% Our study empirically demonstrates that prior LLM-based evaluation studies are insufficient than EX to evaluate text-to-SQL systems, raising the need for a more advanced evaluation paradigm.
\section{Our Proposed Metric: \ours}
\label{sec:method}
\begin{figure}[!ht]
  \includegraphics[width=0.48\textwidth]{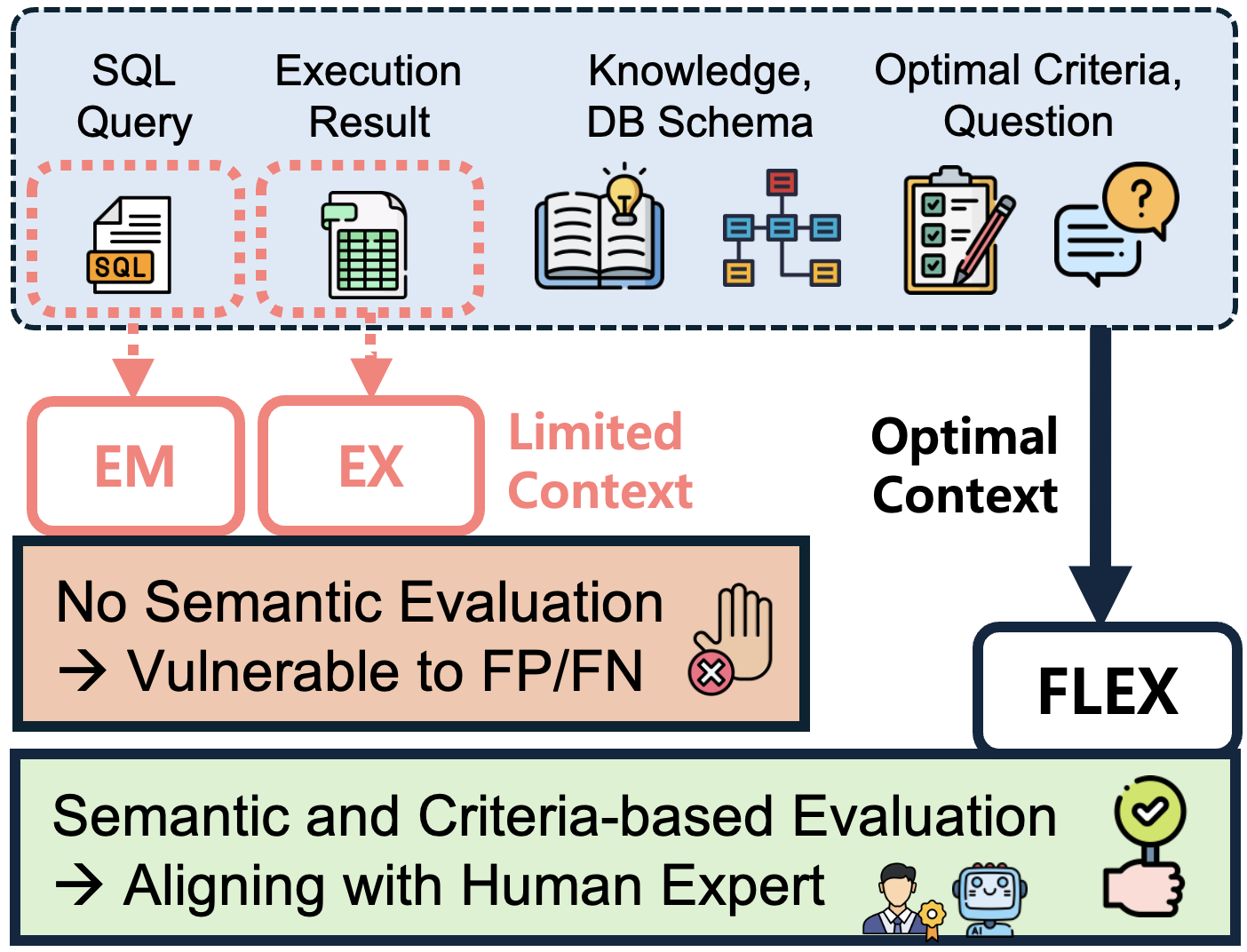}
  \caption{Compared to conventional EM and EX, \ours~evaluates semantic equivalence between question and query based on holistic, contextual information. }
  \label{fig:fig-tp-table}
\end{figure}
This section introduces our proposed evaluation metric \ours~(\textbf{False-Less EXecution}). \ours~addresses the limitations of existing metrics by leveraging LLMs to provide a more accurate and human-like assessment of generated SQL queries.

\subsection{Evaluation Process}
\ours~instructs the LLM with the question, \sqlgen~and optimal contextual information $\mathbb{C}_{FLEX}$, including the question $x$, generated query \sqlgen, ground truth query \sqlgt, execution results \exgt~and \exgen, schema $S$, external knowledge $K$ and two criteria $T_{EQ}$ and $T_{NEQ}$\footnote{Detailed prompts are provided in Appendix~\ref{sec:appendix_prompt_flex}}.
In other words, \ours~prompts LLM to focus on detecting false positives or false negatives after obtaining EX results whether $R_{gen}(x) = R_{gt}(x)$. 
% In other words, here we aim to enhance the context from $\mathbb{C}$ to an improved context $\mathbb{C}_{FLEX}$. By providing the LLM with optimal contextual information—including the question $x$, generated query \sqlgen, ground truth query \sqlgt, execution results \exgt~and \exgen, schema $S$, external knowledge $K$, and two criteria $T{EQ}$ and $T_{NEQ}$\footnote{Detailed prompts are provided in Appendix~\ref{sec:appendix_prompt_flex}}—we enable it to focus on detecting false positives or false negatives after obtaining execution results, specifically by determining whether $R_{gen}(x) = R_{gt}(x)$. This refined context helps guide the LLM in assessing the equivalence between the generated and ground truth execution results.

% This approach is more effective and better aligned with expert judgment than fixed contextual information.
% here we aim to optimize the context between $\mathbb{C}_{EQ}$ to $\mathbb{C}_{NEQ}$, guiding LLM to detect false positives or false negatives through specialized criteria.
% Therefore, we define two scenarios depending on the $R_{gen}(x) = R_{gt}(x)$,
% This approach aligns the LLM judgment better with human experts, \ours~utilizes optimal contextual information between $\mathbb{C}_{EQ}$ and $\mathbb{C}_{NEQ}$ .
% After evaluating all samples in the dataset, we compute the overall accuracy $A$ of the text-to-SQL model as the proportion of queries judged correct by \ours:
% \begin{align} 
%     A = \frac{1}{|Q|}\sum_{q \in Q}\mathbb{I}\left( \mathcal{L}(q) = \text{Correct} \right) 
% \end{align}
\vspace{-5pt}
\begin{equation}
\begin{aligned}
\mathbb{C}_{FLEX} &= \begin{cases}
\mathbb{C}_{EQ}  & \text{if } R_{gen}(x) = R_{gt}(x) \\
\mathbb{C}_{NEQ}  & \text{if } R_{gen}(x) \neq R_{gt}(x)
\end{cases} \\
\mathbb{C}_{base} &= \{x, S, K, Q_{gt}(x), Q_{gen}(x)\}\\
\mathbb{C}_{EQ} &= \mathbb{C}_{base} \cup \{T_{EQ}\}\\
\mathbb{C}_{NEQ} &= \mathbb{C}_{base} \cup \{T_{NEQ}, R_{gt}(x), R_{gen}(x)\}
\end{aligned}
\end{equation}
\subsection{Optimal Context ($\mathbb{C}_{FLEX}$)}

\noindent\textbf{1) Execution Results Match ($\mathbb{C}_{EQ}$})

If $R_{gen}(x) = R_{gt}(x)$, there is a possibility of false positives—queries that produce correct results by coincidence or through incorrect logic. 
To evaluate false positives, 1) $\mathbb{C}_{EQ}$ does not contain \exgt~and \exgen. 
LLM occasionally judges false positives as correct due to the equivalent execution results.
2) $\mathbb{C}_{EQ}$ utilized the following criteria ($T_{EQ}$) to evaluate semantic correctness between the questions and \sqlgen:
\begin{itemize}[noitemsep,topsep=5pt,parsep=2pt,partopsep=0pt,leftmargin=*]
    \item \textbf{Schema Alignment}: Whether the tables and columns used in \sqlgen~align with the intent of question $q$ and are consistent with the schema $S$.
    \item \textbf{Correct Filtering Conditions}: Whether the WHERE clauses in \sqlgen~accurately reflect the conditions specified in the question $x$.
    \item \textbf{Handling of Nullable Columns}: Whether \sqlgen~properly handles cases where the nullable columns in aggregation functions (SUM, COUNT, AVG) of WHERE clauses, as improper handling can lead to unexpected results.
    \item \textbf{Accounting for Multiple Rows}: Whether \sqlgen~correctly handles cases where multiple rows satisfy the query conditions such as one-to-many relation and multiple min/max tuples. 
    \item \textbf{Abusing of Clauses}: Whether clauses like GROUP BY, HAVING, ORDER BY, or DISTINCT are used appropriately, avoiding unnecessary complexity that could alter the intended results.
\end{itemize}

The LLM analyzes differences between \sqlgen~and \sqlgt~to detect logical discrepancies that might not affect the current execution result but could lead to incorrect outcomes if the database state changes.

\noindent\textbf{2) Execution Results Differ ($\mathbb{C}_{NEQ}$)}

Although $R_{gen}(x) \neq R_{gt}(x)$, \sqlgen~may still be correct due to ambiguities in the question or annotation issues with the ground truth query. 
In this scenario, $\mathbb{C}_{EQ}$ contains \exgt, \exgen, and $T_{NEQ}$ which designed to evaluate semantic correctness between the question and \sqlgen~by considering:

\begin{itemize}[noitemsep,topsep=5pt,parsep=2pt,partopsep=0pt,leftmargin=*]
    \item \textbf{Acceptable Output Structure Variations}: Whether differences in the output structure (e.g., column order, extra or missing columns) are acceptable given the way $x$ is phrased.
    \item \textbf{Representation of Values}: Whether differences in value formats (e.g., numerical precision, percentage representation, boolean values as YES/NO) are acceptable for readability and do not alter the meaning.
    \item \textbf{Multiple Answers Available}: Whether $x$ is interpretable in multiple valid ways, each leading to different correct queries.
    \item \textbf{Incorrect Ground Truth}: Whether the ground truth query \sqlgt~is incorrect or suboptimal, while \sqlgen~correctly answers the question.
\end{itemize}
The LLM thus considers the flexibility required due to ambiguous or underspecified questions and assesses whether \sqlgen~is a valid translation of the question. By focusing on semantic correctness and closely aligning with human expert judgments, \ours~provides a more reliable evaluation of text-to-SQL models than traditional execution-based metrics. It effectively mitigates issues with false positives and false negatives, offering a more accurate assessment of model performance.
% After evaluating all samples in the dataset, we compute the overall accuracy $A$ of the text-to-SQL model as the proportion of queries judged correct by \ours:
% \begin{align}
%     A = \frac{1}{|Q|}\sum_{q \in Q}\mathbb{I}(\mathcal{L}(q, SQL_{gen}(q), SQL_{gt}(q), R_{gen}(q), R_{gt}(q), S, K) = Correct)
% \end{align}
% \begin{align} 
%     A = \frac{1}{|Q|}\sum_{q \in Q}\mathbb{I}\left( \mathcal{L}(q) = \text{Correct} \right) 
% \end{align}
% Here, $\Pi(\cdot)$ is the indicator function, which equals 1 if the condition is true (i.e., the LLM judges the generated query as correct) and 0 otherwise. The function $\mathcal{L}(\cdot)$ represents the LLM's judgment process based on the provided inputs. 
\section{Experiments}
To validate the effectiveness of our \ours~approach, we conducted a comprehensive study comparing the agreement between LLM judgments and human evaluations. We tested various state-of-the-art language models to determine which aligns most closely with human judgment.

\subsection{Does FLEX Outperform Other Metrics?}
\begin{figure}
    \centering
    \includegraphics[width=\linewidth]{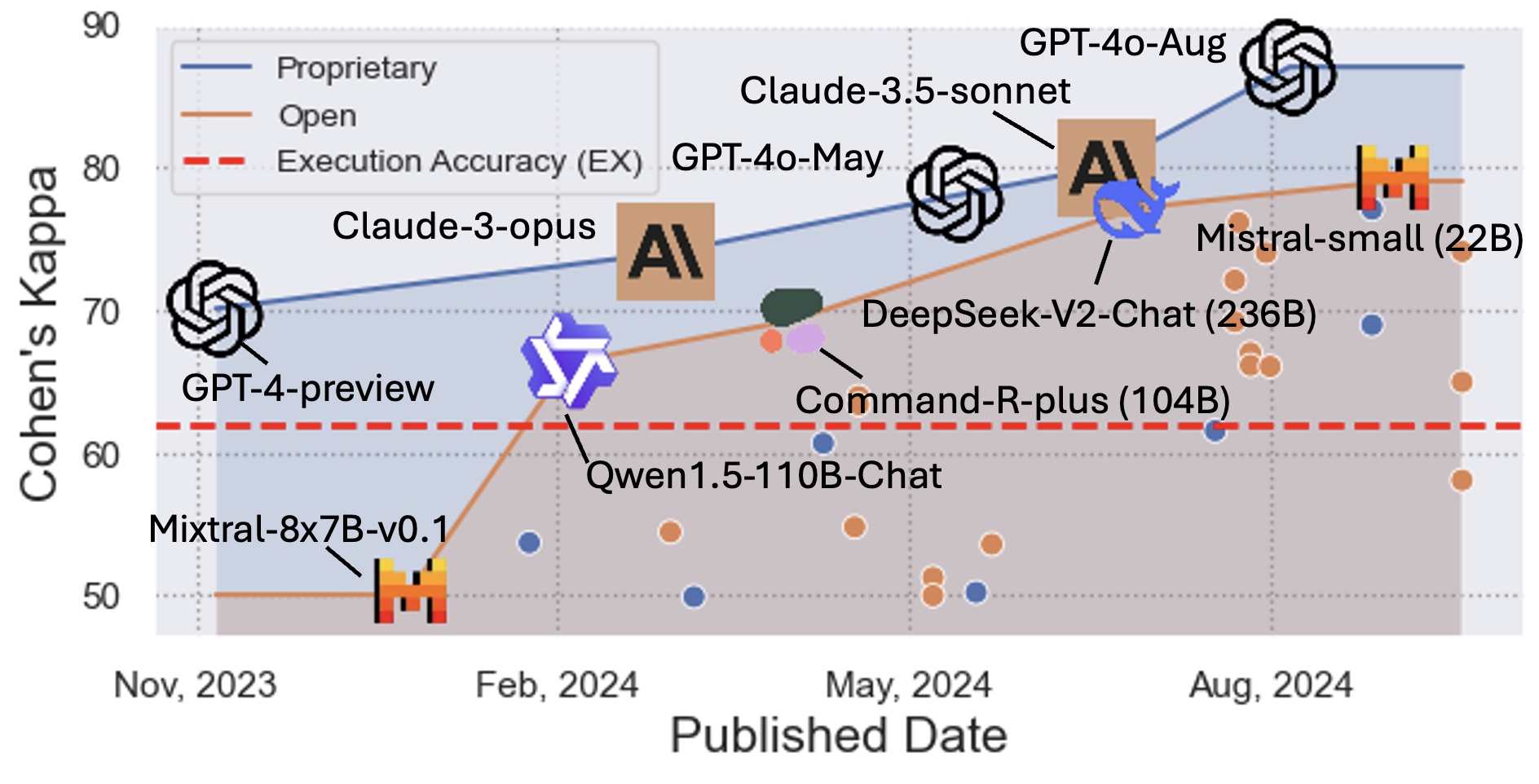}
    \vspace{-20pt}
    \caption{
    Agreements between human evaluation and \ours~across LLM models over time. The \textcolor{red}{red} line shows EX metric agreement. Dots represent other LLMs, illustrating lower agreement than previous SOTA. Details are illustrated in Fig.~\ref{fig:flex_pub_kappa_full}
    % The agreements between human evaluation and \ours~with each LLM model. The  \textcolor{red}{red} line indicates the agreement of EX metric. Dots are other LLMs demonstrating lower agreement than previous SOTA LLMs. Detail results are illustrated in Fig.~\ref{fig:flex_pub_kappa_full}.
    }
    \label{fig:flex_pub_kappa}
\end{figure}
As shown in Fig.~\ref{fig:flex_pub_kappa}, \ours~generally outperforms the traditional EX metric across various LLMs, with GPT-4o achieving the highest agreement with human judgments. Open-source models like Mistral-small-Instruct-2409~\cite{jiang2023mistral} show substantial improvement over the EX metric, indicating that expert-level evaluations are achievable without reliance on proprietary solutions.
LLM performance has improved significantly over time, with more recent models generally demonstrating higher Cohen's Kappa scores. Although open-source models have consistently underperformed proprietary models such as GPT-4o, they are steadily advancing and closing the gap. 
For instance, DeepSeek-V2-Chat~\cite{deepseekai2024deepseekv2strongeconomicalefficient} shows close performance to Claude-3.5-sonnet~\cite{Claude35Sonnet}, a proprietary LLM in its contemporary time. 

% Furthermore, model size is a critical factor in performance improvement as illustrated by models Gemma 2~\cite{gemmateam2024gemmaopenmodelsbased}, Llama 3.1~\cite{dubey2024llama3herdmodels}, and Qwen2.5~\cite{qwen2.5} in Fig.~\ref{fig:flex_pub_kappa_full}. We hypothesize that larger models contain more comprehensive knowledge of SQL and database concepts, allowing for more nuanced and accurate evaluations. This is consistent with scaling law in knowledge capacity~\cite{allen2024physics}.
\ours~also offers substantial efficiency gains in both time and cost. GPT-4o evaluated the BIRD dev set (1,534 instances) in under 20 minutes at approximately \$6, 
whereas volunteers required an average of 2 hours for 200 instances. These results demonstrate \ours's viability as an efficient (42x faster) and cost-effective alternative to both EX and human evaluation for text-to-SQL tasks.

\subsection{What Factors are Beneficial?}
% \begin{table}
% \centering
% \caption{Comparison of human agreement results depending on execution results.}
% \label{tab:tab_flex_ablation_cond}
% \adjustbox{max width=\linewidth}{
% \begin{tabular}{cc|cccc} 
% \hline
% $\mathbf{R_{gen}(x)}$ & $\mathbf{R_{gt}(x)}$  & \textbf{Kappa} & \textbf{Acc}  & \textbf{EQ} & \textbf{NEQ}  \\ 
% \hline
% \multicolumn{2}{c|}{$\mathbb{C}_{FLEX}$}   & \textbf{87.04} & \textbf{93.5} & \textbf{88} & \textbf{99}   \\ 
% \hdashline
% $\times$                & $\times$   & 82.06          & 91.0          & \textbf{88} & 94            \\
% $\circ$                & $\circ$   & 81.08          & 90.5          & 82          & \textbf{99}   \\
% \hline
% \end{tabular}
% }
% \end{table}
\begin{table}[ht]
\centering
\caption{Comparison of human agreement results depending on execution results.}
\label{tab:tab_flex_ablation_cond}
\normalsize
\setlength{\tabcolsep}{3pt}
\adjustbox{max width=0.95\linewidth}{
\begin{tabular}{cccc|cccc} 
\hline
\multicolumn{2}{c}{$\mathbb{C}_{EQ}$} & \multicolumn{2}{c|}{$\mathbb{C}_{NEQ}$}  & \multirow{2}{*}{\textbf{Kappa}} & \multirow{2}{*}{\textbf{Acc}}  & \multirow{2}{*}{\textbf{EQ}} & \multirow{2}{*}{\textbf{NEQ}}  \\ 
$\mathbf{R_{gen}(x)}$ & $\mathbf{R_{gt}(x)}$ & $\mathbf{R_{gen}(x)}$ & $\mathbf{R_{gt}(x)}$ & & & & \\ 
\hline
$\times$ & $\times$ &$\circ$&$\circ$&\textbf{87.04} & \textbf{93.5} & \textbf{88} & \textbf{99}   \\  
\hdashline
$\times$ & $\times$ & $\times$ & $\times$ & 82.06 & 91.0 & \textbf{88} & 94       \\ 
$\circ$ & $\circ$ & $\circ$ & $\circ$ & 81.08          & 90.5          & 82          & \textbf{99}   \\ 
\hline
\end{tabular}
}
\end{table}
% \begin{table}
% \centering
% \caption{Comparison of human agreement results using different contextual information. The optimal context ($\mathbb{C}_{FLEX}$) shows the highest agreement. The flexible inclusion of execution results shows the highest agreement.}
% \label{tab:tab_flex_ablation_cond}
% \adjustbox{max width=\linewidth}{
% \begin{tabular}{@{\extracolsep{4pt}}cccc@{}|cccc} 
% \hline
% \multicolumn{2}{c}{$\mathbb{C}_{EQ}$} & \multicolumn{2}{c|}{$\mathbb{C}_{NEQ}$}  & \multirow{2}{*}{\textbf{Kappa}} & \multirow{2}{*}{\textbf{Acc}}  & \multirow{2}{*}{\textbf{EQ}} & \multirow{2}{*}{\textbf{NEQ}}  \\ 
% \noalign{\global\setlength{\arrayrulewidth}{0.5mm}}
% \cline{1-2} \cline{3-4}
% $\mathbf{R_{gen}(x)}$ & $\mathbf{R_{gt}(x)}$ & $\mathbf{R_{gen}(x)}$ & $\mathbf{R_{gt}(x)}$ & & & & \\ 
% \hline
% $\times$ & $\times$ &$\circ$&$\circ$&\textbf{87.04} & \textbf{93.5} & \textbf{88} & \textbf{99}   \\  
% \hdashline
% $\times$ & $\times$ & $\times$ & $\times$ &\textbf{82.06} & \textbf{91.0} & \textbf{88} & 94       \\ 
% $\circ$ & $\circ$ & $\circ$ & $\circ$ & 81.08          & 90.5          & 82          & \textbf{99}   \\ 
% \hline
% \end{tabular}
% }
% \end{table}

\begin{table}[ht]
\centering
\caption{Comparison of human agreement results using different contextual information.}
\label{tab:tab_flex_ablation_base}
\small
\adjustbox{max width=\linewidth}{
\begin{tabular}{l|cccc} 
\hline
\textbf{Ablation Settings} & \textbf{Kappa} & \textbf{Acc} & \textbf{EQ} & \textbf{NEQ}  \\ 
\hline
\textit{w/o Question}               & 80.10          & 90.0         & 84  & 96            \\
\textit{w/o~Knowledge}              & 79.09          & 89.5         & 82          & 97    \\
\textit{w/o~Criteria}               & 74.08          & 87.0         & 81          & 93            \\
\textit{w/o~Ground Truth}           & 29.36          & 64.0         & 72          & 56            \\
\hline
\end{tabular}
}
\end{table}
We conducted an ablation study to validate the efficacy of the optimal context $\mathbb{C}_{FLEX}$ and each contextual information. As shown in Table~\ref{tab:tab_flex_ablation_cond}, including execution results only in $\mathbb{C}_{NEQ}$ yields the best performance across all metrics. 
When execution results are included in both $\mathbb{C}_{EQ}$ and $\mathbb{C}_{NEQ}$, performance is strong in the \textit{Equivalent set} (true positives + false positives) but declines in the \textit{Not equivalent set} (true positives + false negatives). 
Conversely, excluding execution results from both contexts shows the opposite trend.
The inclusion of execution results impacts the model's ability to detect errors. Equivalent execution results may hinder false positive detection even if the generated query has incorrect logic. Conversely, for non-equivalent queries, including execution results helps reduce false negatives, as the model can directly compare minor differences in execution results. Therefore, the optimal context selection approach enhances evaluation for \sqlgen, aligning it more closely with human expert assessments.
% We conducted an ablation study to validate the efficacy of optimal context $\mathbb{C}_{FLEX}$ and each contextual information.
% As shown in Table~\ref{tab:tab_flex_ablation_cond}, including the execution results only in $\mathbb{C}_{NEQ}$ show best performances across every metric. 
% Inclusion and exclusion of the execution result in both $\mathbb{C}_{EQ}$ and $\mathbb{C}_{NEQ}$ lead to performance drop in the \textit{Equivalent set} and \textit{Not equivalent set}, respectively.
% Including execution results influences performance based on the equivalence of the generated query's execution result to the ground truth. Equivalent execution results may hinder false positive detection even if the generated query has incorrect logic. 
% Conversely, for non-equivalent queries, including execution results, a false negative evaluation is beneficial as the LLM can directly compare minor differences.
% Therefore, the optimal context selection approach benefits \sqlgen~ evaluation, aligning it more closely with human experts.

Table~\ref{tab:tab_flex_ablation_base} shows the impact of removing different components from the evaluation context. Removing the ground truth query and results is the most significant factor, causing a substantial performance drop across all metrics. This underscores the importance of a reliable reference point in LLM-based text-to-SQL evaluations.
Removing the natural language question or external knowledge also decreases notable performance, emphasizing their role in providing a comprehensive evaluation context.
The removal of evaluation criteria also impacts performance, highlighting the importance of well-defined guidelines.
This ablation study demonstrates that \ours's performance stems from the synergy of all its components, each playing a vital role in achieving high-quality text-to-SQL evaluation.
\section{Leaderboard Re-evaluation}
In this section, we evaluated the published result in the Spider and BIRD benchmark using robust capabilities of the \ours. 
Three key findings were observed to demonstrate crucial limitations of current text-to-SQL evaluation.
We first introduce experimental settings and demonstrate three crucial limitations.

\subsection{Experiment Setup}
For sections~\ref{sec:Finding1} and \ref{sec:Finding3}, we conducted experiments using all of the publicly available 50 models on both the Spider and BIRD benchmarks.
For section~\ref{sec:Finding2}, we examine error categories on the top ten models in BIRD and Spider benchmarks.
Our study utilizes the dev set of the BIRD benchmark because it has not disclosed their test set.
For a fair comparison, we employed a dev set of the Spider benchmark; most studies published their prediction results of the dev set.
GPT-4o was employed for the query evaluation as an LLM judge, which shows an outperforming agreement in our human evaluation. 

% \subsection{Findings 1: Leaderboard Shifts Significantly}
\subsection{Findings 1: Leaderboard Shifts}
\label{sec:Finding1}
The change in the performance of text-to-SQL models in our re-evaluation leads to a significant shift in the leaderboard.
We found that the EX metric mostly underestimates the models.
As illustrated in Fig.~\ref{fig:scatter-ex-flex-spider} and \ref{fig:scatter-ex-flex-bird}, the average increases from EX to \ours~are $+2.63$ and $+2.6$ in Spider and BIRD benchmarks, respectively.
This result causes an average rank change of about 1.7 and 2.2 in both benchmarks. 
Specifically, the maximum rank changes are as large as five positions in BIRD and eight in Spider.
These findings demonstrate that the EX metric obscures models' true performance, leading to inaccurate leaderboard rankings.

\begin{figure}
\includegraphics[width=\linewidth]{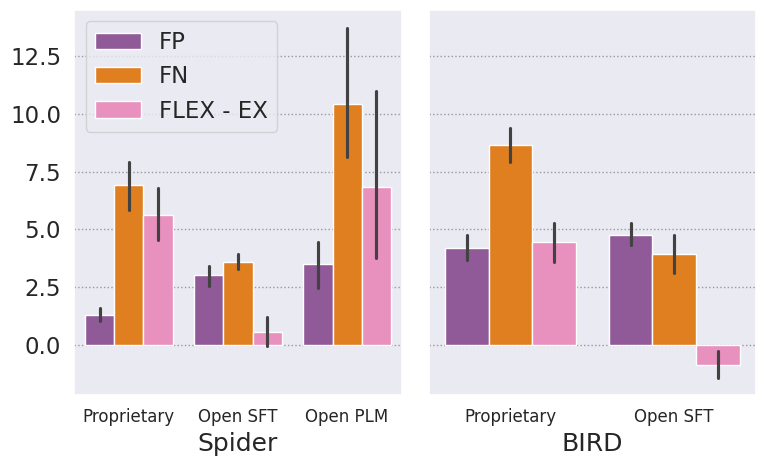}
\caption{Average model performances and error ratios across different model types.}
\label{fig:modeltype-barplot-flex-ex-fp-fn}
\end{figure}
The extent of these shifts varies across different model types: Proprietary, open pre-trained language models (open PLMs), and open language models with supervised fine-tuning (open SFT). 
As described in Fig.~\ref{fig:modeltype-barplot-flex-ex-fp-fn}, proprietary models and open PLMs show significant improvements when re-evaluated using the ~\ours, whereas open SFT models display only marginal changes. 
A closer examination of the error types explains this difference. Open SFT models have a lower false negative ratio than proprietary and open PLM models. In contrast, all model types show relatively similar false positive ratios. 

We hypothesize that this phenomenon can be attributed to differences in the training approaches. 
The supervised fine-tuning process enables SFT models to predict SQL queries structurally similar to those in the training dataset \cite{li2024dawn}. 
This similarity results in execution results closely matching those of the ground-truth queries, reducing false negatives. 
In contrast, proprietary and open PLM models are more likely to generate more diverse SQL queries. 
This diversity increases the likelihood of structural differences between the execution results of the predicted and ground-truth queries. 
Consequently, these models are more susceptible to false negatives under the EX metric, which \ours~can correct.

\subsection{Findings 2: Models are Underestimated due to Annotation Quality}
\label{sec:Finding2}
\begin{figure}[!ht]
    \centering
    \includegraphics[width=0.95\linewidth]{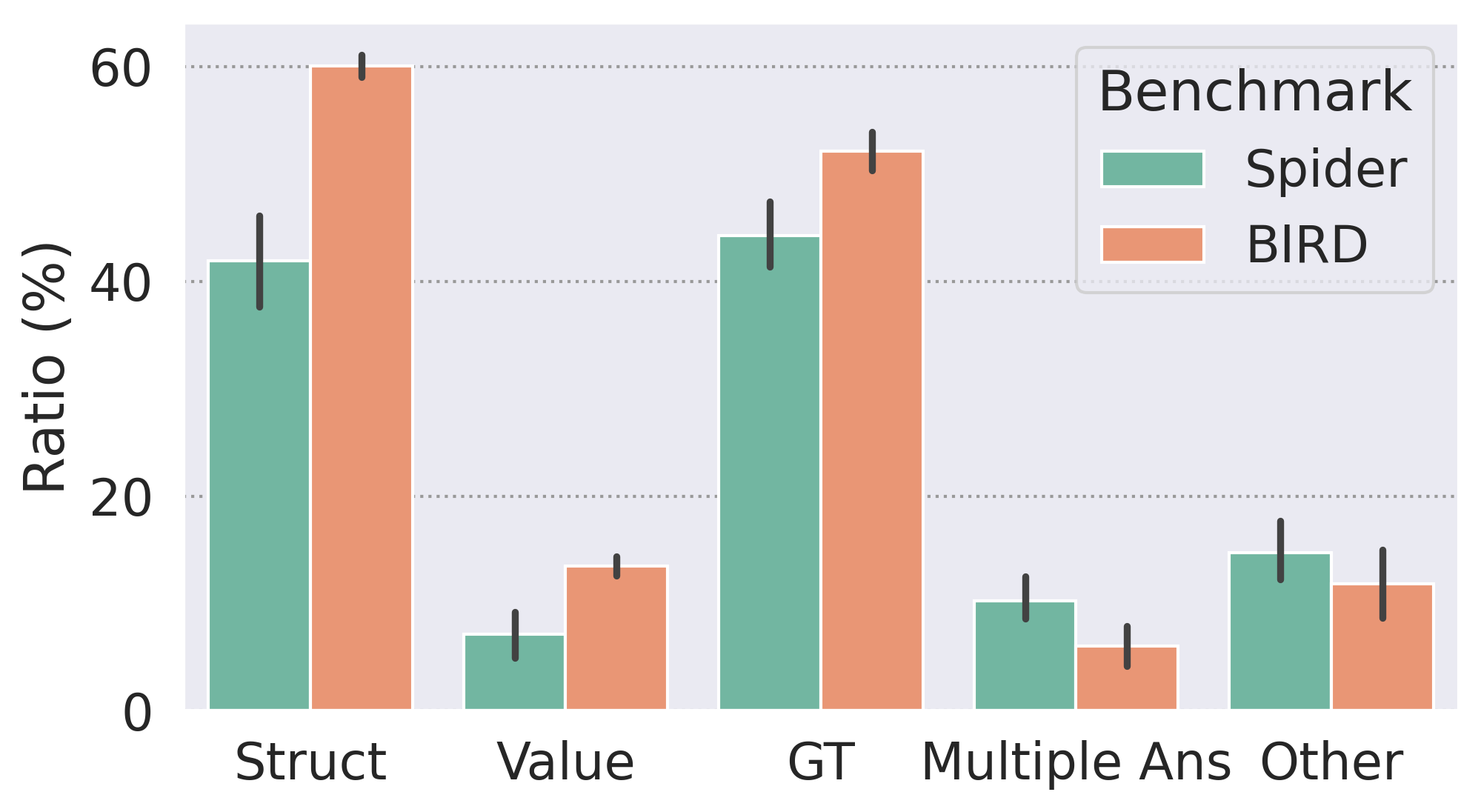}
    \caption{Categorized result of FN ratios in top 10 models. Struct denotes an acceptable output structure variation, Value denotes a different representation of value, GT denotes incorrect ground truth, and Multiple Ans denotes multiple answers available.}
    \label{fig:cat_top10_fn}
\end{figure}
We analyze the causes of false negatives, which lead to underestimation of text-to-SQL models' performance, by leveraging a categorized error summary from our framework\footnote{Detailed process is provided in appendix~\ref{sec:appendix_error_cat}}.
As illustrated in Fig.~\ref{fig:cat_top10_fn}, the primary causes of false negatives are different output structures (Struct) and incorrect ground truth (GT). The inherent ambiguity in questions allows for several possible answers, leading to many false negatives due to different output structures (58.13\% in BIRD). Incorrect ground truth SQL queries also account for a significant portion of false negatives (54.54\% in BIRD). These results support the observation of previous works~\cite{wang2023know, wretblad2024understanding} and show that this limitation remains unaddressed even in state-of-the-art models.

\subsection{Findings 3: Models are Overestimated in Difficult Questions}
\label{sec:Finding3}
\begin{figure}[!ht]
\includegraphics[width=\linewidth]{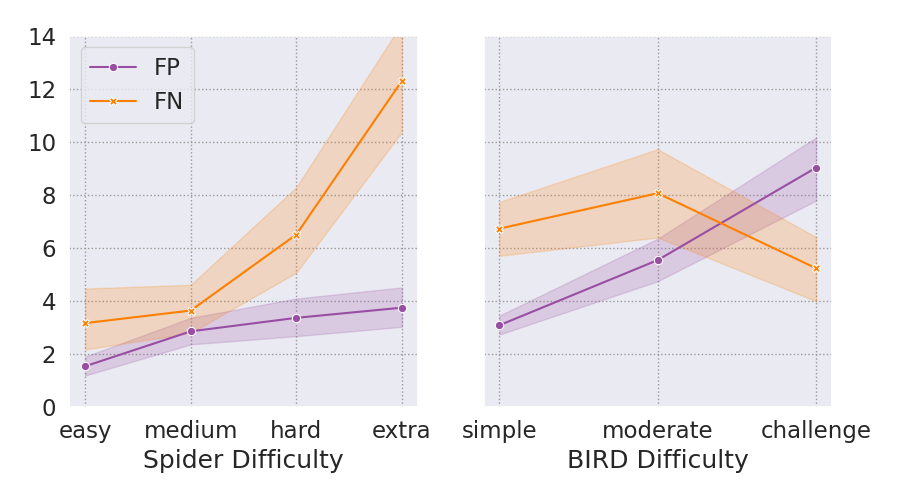}
\vspace{-20pt}
\caption{The models' average FP and FN ratios across difficulty levels. The y-axis denotes the percentage~(\%). In the Spider, text-to-SQL models tend to be underestimated as the questions become more difficult. Conversely, in the BIRD, models are overestimated with more challenging questions.}
\label{fig:hardness-flex-ex-fp-fn-line}
\end{figure}

Our study reveals that text-to-SQL models demonstrate lower \ours~score than EX in BIRD's challenging questions while higher in easier ones. This discrepancy can be elucidated through an analysis of false positive and false negative ratios illustrated in Fig.~\ref{fig:hardness-flex-ex-fp-fn-line}. In Spider, considered less complex than BIRD, the false positive ratio is consistently lower than the false negative ratio for text-to-SQL models, indicating that text-to-SQL models show higher \ours~than EX. This trend persists for simple and moderate questions in the BIRD. However, in challenging questions of BIRD, the false positive ratio exceeds the false negative ratio, leading to a decrease in the model's performance. These results underscore the significant impact of false positives and false negatives in text-to-SQL evaluation.
\section{Conclusion}
In this paper, we found the limitations in text-to-SQL evaluation, the vulnerability of the EX metric to false positives and negatives, which can result in inaccurate model assessments. Although LLM-based evaluations have emerged as alternatives to human evaluators, they still reveal several limitations. To overcome this, we introduce a novel metric, \ours~\textbf{(False-Less EXecution)}. It utilizes the comprehensive context and elaborate criteria based on the advanced language understanding of LLMs to achieve more expert-level evaluations. Our comprehensive experiments show the effectiveness of our proposed metric and provide several empirical findings to promote future studies on text-to-SQL evaluation.

\section{Limitations}
While our \ours~evaluation approach offers significant improvements over traditional metrics like Execution Accuracy (EX), there are some limitations to note: \ours~has a strong agreement; Cohen's kappa of 87.04 is imperfect and could be further improved through advanced reasoning~\cite{wang2022selfconsistency, madaan2024self} or fine-tuning~\cite{wang2023pandalm, zhu2023judgelm} of the LLM evaluator.
Our evaluation is currently limited to the Spider and BIRD benchmarks, necessitating further testing on other text-to-SQL datasets and real-world enterprise databases to validate \ours's effectiveness fully. \ours's reliance on LLMs makes it computationally intensive and more time-consuming than simple execution-based metrics, which may limit scalability for million-scale evaluations. Moreover, proprietary LLMs pose reproducibility risks, as providers can change or deprecate models without notice. While \ours~provides a more comprehensive assessment, it does not eliminate the need for human review, especially in critical applications. Addressing these limitations presents opportunities for future work to refine further and expand LLM-based evaluation approaches for text-to-SQL systems.

\bibliography{custom}
\clearpage
\appendix

\section{Experiment Setting}
\label{sec:appendix}

\subsection{Frameworks and Hardware}
We utilized NVIDIA A6000x4 GPUs with Huggingface transformers~\cite{wolf-etal-2020-transformers} and VLLM~\cite{kwon2023efficientvllm} for generating outputs from open LLMs. To evaluate open models larger than 30B, we utilize the \href{https://huggingface.co/docs/api-inference/index}{Huggingface Inference API}, \href{https://www.together.ai/}{TogetherAI Inference}, and \href{https://aimlapi.com/}{AI/ML API}.

\subsection{Generation Hyperparameter}
We set the temperature to 0 in every experiment to ensure consistent results for each attempt and the maximum number of tokens to 2,048.

\subsection{Execution Results in the LLM Prompt}
The execution results in the LLM prompt were converted to a markdown table. However, the SQL query can produce extensive rows, increasing the prompt length and API costs. To reduce the overhead, we truncated the middle of the table with a mark as `...' and left 50 heads and tails for more than 100 rows. The shape of the table is appended to the markdown table to help LLM's judgment. Some TEXT-type columns can also be longer. After the 50 characters, we truncated them and appended `\textit{... k chars}', where k indicates an original number of characters.

\renewcommand{\arraystretch}{1.0}

\begin{table*}
\centering
\caption{Simplified names for DAIL-SQL model results}
\label{tab:dail-sql-results}
\small
\begin{adjustbox}{max width=\textwidth}
\begin{tabular}{|l|l|} 
\hline
\textbf{Simplified Name}       & \textbf{Original File Name}                                                                                                                           \\ 
\hline
DAIL\_SQL\_7-SHOT\_TH-0.85\_MP & \begin{tabular}[c]{@{}l@{}}BIRD\_WITH\_EVIDENCE-TEST\_SQL\_7-SHOT\_EUCDISMASKPRESKLSIMTHR\_QA-\\EXAMPLE\_CTX-150\_ANS-4096\_TH-0.85.txt\end{tabular}  \\ 
\hline
DAIL\_SQL\_7-SHOT\_TH-0.8\_MP  & \begin{tabular}[c]{@{}l@{}}BIRD\_WITH\_EVIDENCE-TEST\_SQL\_7-SHOT\_EUCDISMASKPRESKLSIMTHR\_QA-\\EXAMPLE\_CTX-150\_ANS-4096\_TH-0.8.txt\end{tabular}   \\ 
\hline
DAIL\_SQL\_9-SHOT\_MP          & \begin{tabular}[c]{@{}l@{}}BIRD\_WITH\_EVIDENCE-TEST\_SQL\_9-SHOT\_EUCDISMASKPRESKLSIMTHR\_QA-\\EXAMPLE\_CTX-150\_ANS-4096.txt\end{tabular}           \\ 
\hline
DAIL\_SQL\_9-SHOT\_QM          & \begin{tabular}[c]{@{}l@{}}BIRD\_WITH\_EVIDENCE-TEST\_SQL\_9-SHOT\_EUCDISQUESTIONMASK\_QA-\\EXAMPLE\_CTX-150\_ANS-4096.txt\end{tabular}               \\ 
\hline
DAIL\_SQL                      & QUESTIONMASK\_GPT-4\_for\_bird.txt                                                                                                                    \\
\hline
\end{tabular}
\end{adjustbox}
\end{table*}

\subsection{Source of Prediction Results}
The prediction results were obtained from various sources, including NL2SQL360~\cite{li2024dawn} frameworks, published studies, and reproduced predictions. Table~\ref{tab:prediction_sources} provides an overview of these sources.
In DAIL-SQL~\cite{gao2023text}, we simplified model results in their repository\footnote{\url{https://github.com/BeachWang/DAIL-SQL/tree/main/results}} for readability, as shown in Table~\ref{tab:dail-sql-results}.

% Table for prediction sources
\begin{table}[htbp]
\centering
\caption{Overview of prediction sources}
\begin{tabularx}{\linewidth}{lX}
\toprule
\textbf{Category} & \textbf{Frameworks / Studies} \\
\midrule
NL2SQL360 & DIN-SQL~\cite{pourreza2023dinsql}, C3\_SQL~\cite{dong2023c3}, CodeS~\cite{li2024codes}, RESDSQL~\cite{li2023resdsqldecouplingschemalinking}, Graphix~\cite{li2023graphix}, DeepSeek Coder~\cite{guo2024deepseekcoderlargelanguagemodel}, Llama-2~\cite{touvron2023llama2openfoundation}, Llama-3~\cite{dubey2024llama3herdmodels}, Code Llama~\cite{roziere2024codellamaopenfoundation}, StarCoder~\cite{li2023starcodersourceyou} \\
Published & TA-ACL~\cite{qu2024generationalignitnovel}, DAIL-SQL~\cite{gao2023text}, BIRD's baseline (GPT-4)~\cite{openai2024gpt4technicalreport} \\
Reproduced & DTS-SQL (GPT-4o)~\cite{pourreza2024dtssqldecomposedtexttosqlsmall}, CHESS (GPT-4o-mini)~\cite{talaei2024CHESS} \\
\bottomrule
\end{tabularx}
\label{tab:prediction_sources}
\end{table}

\subsection{Human Annotators}
Three of the authors are annotators. They have taken database classes at the university and have more than three years of experience operating the database system.

\section{Additional Experiment Result}

\subsection{FLEX Human Agreement across Various LLMs}
\label{sec:flex_pub_kappa_full}

\begin{figure}
    \centering
    \includegraphics[width=\linewidth]{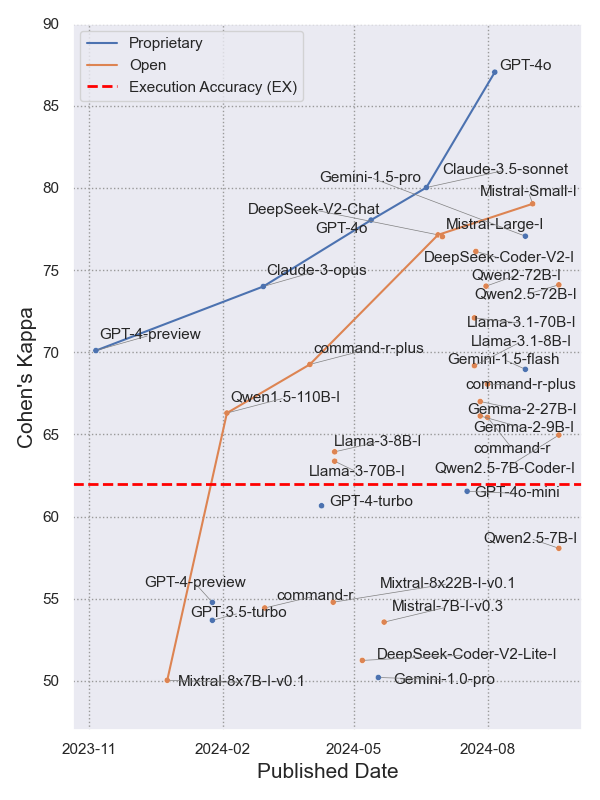}
    \caption{Detailed results of Agreement between Human Evaluation and FLEX across LLM models. The \textcolor{red}{red} line indicates the agreement of EX metric. ``-I'' denotes an instruction-tuned model}
    \label{fig:flex_pub_kappa_full}
\end{figure}

Fig.~\ref{fig:flex_pub_kappa_full} shows the human-LLM agreement results. We employed the following cutting-edge LLMs to evaluate human-LLM agreements. For proprietary models, we employed OpenAI's GPT models~\cite{achiam2023gpt,openai2024gpt4technicalreport}, Anthropic's Claude models~\cite{Claude35Sonnet}, and Google's Gemini models~\cite{geminiteam2024gemini15unlockingmultimodal}.

For open LMs, we employed Meta's Llama-3 and Llama-3.1 models~\cite{dubey2024llama3herdmodels}, MistralAI's models~\cite{jiang2023mistral, mixtral8x7B}, Google's Gemma-2 models~\cite{gemmateam2024gemmaopenmodelsbased}, Qwen-1.5~\cite{qwen}, Qwen-2~\cite{qwen2} and Qwen-2.5 models~\cite{qwen2.5} DeepSeek AI's DeepSeek-V2~\cite{deepseekai2024deepseekv2strongeconomicalefficient} and DeepSeek-Coder~\cite{guo2024deepseekcoderlargelanguagemodel}, Cohere's Command-R~\cite{cohere_for_ai_2024}.

% Interestingly, the model size and performance relationship is not strictly linear across all model families. As shown in Fig.~\ref{fig:flex_pub_kappa_full}, while larger models often perform better in open models, the Mistral models present an exception to this trend. In the Mistral~\cite{jiang2023mistral} models, we observe that Mistral-Small outperforms its larger counterparts. This variance suggests that factors beyond mere model size play crucial roles in an LLM's ability to evaluate SQL queries accurately.

\subsection{Full FLEX results}
\begin{figure*}[!ht]
    \centering
    
    \begin{subfigure}{0.48\textwidth}
        \includegraphics[width=\textwidth]{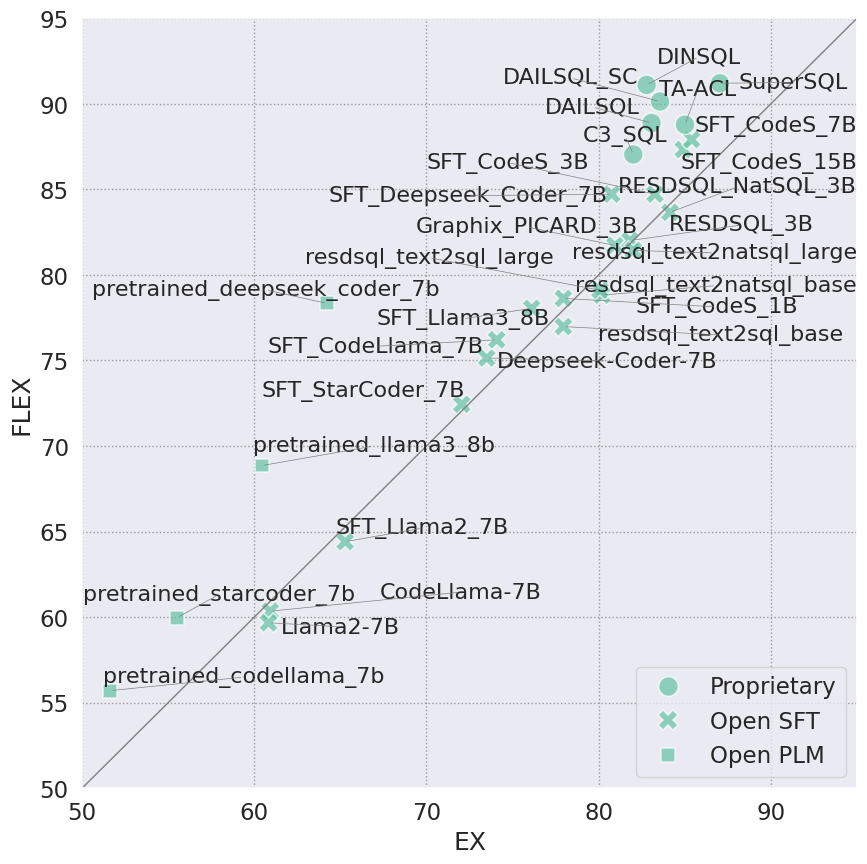}
        \caption{EX vs. \ours~metrics on Spider}
        \label{fig:scatter-ex-flex-spider}
    \end{subfigure}
    \begin{subfigure}{0.48\textwidth}
        \includegraphics[width=\textwidth]{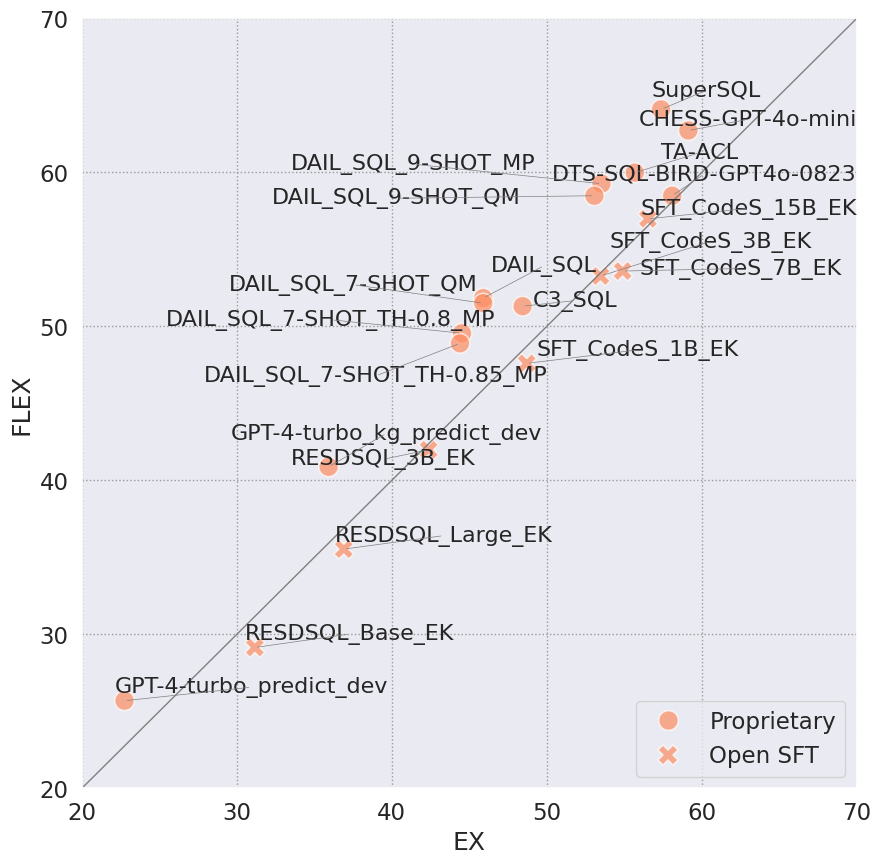}
        \caption{EX vs. \ours~metrics on BIRD}
        \label{fig:scatter-ex-flex-bird}
    \end{subfigure}
    \caption{Comparison between EX and FLEX metrics across two text-to-SQL benchmarks. The gray identity line denotes equivalent scores.}
    \label{fig:scatter-ex-flex}
\end{figure*}

Fig.~\ref{fig:scatter-ex-flex-spider} and \ref{fig:scatter-ex-flex-bird} demonstrate the score comparison between EX and \ours. The gray identity line shows an equivalent score. The full leaderboard of BIRD-dev and Spider-dev is described in table~\ref{tab:tab_full_leaderboard_spider} and \ref{tab:tab_full_leaderboard_bird}, respectively.

\begin{table}[]
\centering
\caption{Spider full result}
\label{tab:tab_full_leaderboard_spider}
\adjustbox{max width=\columnwidth}{
\begin{tabular}{lllll}
\toprule
Rank & Model & FLEX & EX & $\Delta$ \\
\midrule
\phantom{0}1\phantom{0}(-) & SuperSQL & 91.20 & 87.04 & $+4.16$ \\
\phantom{0}2\phantom{0}(\textcolor{red}{↑7}) & DINSQL & 91.10 & 82.79 & $+8.32$ \\
\phantom{0}3\phantom{0}(\textcolor{red}{↑3}) & DAILSQL\_SC & 90.14 & 83.56 & $+6.58$ \\
\phantom{0}4\phantom{0}(\textcolor{red}{↑4}) & DAILSQL & 88.88 & 83.08 & $+5.80$ \\
\phantom{0}5\phantom{0}(\textcolor{blue}{↓2}) & TA-ACL & 88.78 & 85.01 & $+3.77$ \\
\phantom{0}6\phantom{0}(\textcolor{blue}{↓4}) & SFT\_CodeS\_7B & 87.91 & 85.40 & $+2.51$ \\
\phantom{0}7\phantom{0}(\textcolor{blue}{↓3}) & SFT\_CodeS\_15B & 87.33 & 84.91 & $+2.42$ \\
\phantom{0}8\phantom{0}(\textcolor{red}{↑2}) & C3\_SQL & 87.04 & 82.01 & $+5.03$ \\
\phantom{0}9\phantom{0}(\textcolor{red}{↑5}) & SFT\_Deepseek\_Coder\_7B & 84.72 & 80.75 & $+3.97$ \\
10\phantom{0}(\textcolor{blue}{↓2}) & SFT\_CodeS\_3B & 84.72 & 83.27 & $+1.45$ \\
11\phantom{0}(\textcolor{blue}{↓6}) & RESDSQL\_NatSQL\_3B & 83.66 & 84.14 & $-0.48$ \\
12\phantom{0}(-) & RESDSQL\_3B & 82.01 & 81.82 & $+0.19$ \\
13\phantom{0}(-) & Graphix\_PICARD\_3B & 81.72 & 80.95 & $+0.77$ \\
14\phantom{0}(\textcolor{blue}{↓4}) & resdsql\_text2natsql\_large & 81.43 & 82.01 & $-0.58$ \\
15\phantom{0}(\textcolor{red}{↑1}) & resdsql\_text2sql\_large & 79.11 & 80.08 & $-0.97$ \\
16\phantom{0}(\textcolor{blue}{↓1}) & resdsql\_text2natsql\_base & 78.82 & 80.17 & $-1.35$ \\
17\phantom{0}(-) & SFT\_CodeS\_1B & 78.63 & 77.95 & $+0.68$ \\
18\phantom{0}(\textcolor{red}{↑6}) & pretrained\_deepseek\_coder\_7b & 78.34 & 64.22 & $+14.12$ \\
19\phantom{0}(-) & SFT\_Llama3\_8B & 78.05 & 76.11 & $+1.93$ \\
20\phantom{0}(\textcolor{blue}{↓3}) & resdsql\_text2sql\_base & 76.98 & 77.95 & $-0.97$ \\
21\phantom{0}(\textcolor{blue}{↓1}) & SFT\_CodeLlama\_7B & 76.21 & 74.08 & $+2.13$ \\
22\phantom{0}(\textcolor{blue}{↓1}) & Deepseek-Coder-7B & 75.15 & 73.50 & $+1.64$ \\
23\phantom{0}(\textcolor{blue}{↓1}) & SFT\_StarCoder\_7B & 72.44 & 72.05 & $+0.39$ \\
24\phantom{0}(\textcolor{red}{↑3}) & pretrained\_llama3\_8b & 68.86 & 60.44 & $+8.41$ \\
25\phantom{0}(\textcolor{blue}{↓2}) & SFT\_Llama2\_7B & 64.41 & 65.28 & $-0.87$ \\
26\phantom{0}(\textcolor{blue}{↓1}) & CodeLlama-7B & 60.35 & 60.93 & $-0.58$ \\
27\phantom{0}(\textcolor{red}{↑1}) & pretrained\_starcoder\_7b & 59.96 & 55.51 & $+4.45$ \\
28\phantom{0}(\textcolor{blue}{↓2}) & Llama2-7B & 59.67 & 60.83 & $-1.16$ \\
29\phantom{0}(-) & pretrained\_codellama\_7b & 55.71 & 51.64 & $+4.06$ \\
30\phantom{0}(-) & pretrained\_llama2\_7b & 24.08 & 20.99 & $+3.09$ \\
\bottomrule
\end{tabular}
}
\end{table}

\begin{table}[]
\centering
\caption{BIRD full result}
\label{tab:tab_full_leaderboard_bird}
\adjustbox{max width=\columnwidth}{
\begin{tabular}{lllll}
\toprule
Rank & Model & FLEX & EX & $\Delta$ \\
\midrule
\phantom{0}1\phantom{0}(\textcolor{red}{↑2}) & SuperSQL & 64.08 & 57.37 & $+6.71$ \\
\phantom{0}2\phantom{0}(\textcolor{blue}{↓1}) & CHESS-GPT-4o-mini & 62.71 & 59.13 & $+3.59$ \\
\phantom{0}3\phantom{0}(\textcolor{red}{↑2}) & TA-ACL & 59.97 & 55.67 & $+4.30$ \\
\phantom{0}4\phantom{0}(\textcolor{red}{↑3}) & DAIL\_SQL\_9-SHOT\_MP & 59.26 & 53.52 & $+5.74$ \\
\phantom{0}5\phantom{0}(\textcolor{red}{↑4}) & DAIL\_SQL\_9-SHOT\_QM & 58.47 & 53.06 & $+5.41$ \\
\phantom{0}5\phantom{0}(\textcolor{blue}{↓3}) & DTS-SQL-BIRD-GPT4o-0823 & 58.47 & 58.08 & $+0.39$ \\
\phantom{0}7\phantom{0}(\textcolor{blue}{↓3}) & SFT\_CodeS\_15B\_EK & 56.98 & 56.52 & $+0.46$ \\
\phantom{0}8\phantom{0}(\textcolor{blue}{↓2}) & SFT\_CodeS\_7B\_EK & 53.59 & 54.89 & $-1.30$ \\
\phantom{0}9\phantom{0}(\textcolor{blue}{↓1}) & SFT\_CodeS\_3B\_EK & 53.26 & 53.46 & $-0.20$ \\
10\phantom{0}(\textcolor{red}{↑2}) & DAIL\_SQL & 51.83 & 45.89 & $+5.93$ \\
11\phantom{0}(\textcolor{red}{↑1}) & DAIL\_SQL\_7-SHOT\_QM & 51.50 & 45.89 & $+5.61$ \\
12\phantom{0}(\textcolor{blue}{↓1}) & C3\_SQL & 51.30 & 48.44 & $+2.87$ \\
13\phantom{0}(\textcolor{red}{↑1}) & DAIL\_SQL\_7-SHOT\_TH-0.8\_MP & 49.54 & 44.52 & $+5.02$ \\
14\phantom{0}(\textcolor{red}{↑1}) & DAIL\_SQL\_7-SHOT\_TH-0.85\_MP & 48.89 & 44.39 & $+4.50$ \\
15\phantom{0}(\textcolor{blue}{↓5}) & SFT\_CodeS\_1B\_EK & 47.59 & 48.70 & $-1.11$ \\
16\phantom{0}(-) & RESDSQL\_3B\_EK & 41.98 & 42.37 & $-0.39$ \\
17\phantom{0}(\textcolor{red}{↑1}) & GPT-4-turbo\_kg\_predict\_dev & 40.87 & 35.92 & $+4.95$ \\
18\phantom{0}(\textcolor{blue}{↓1}) & RESDSQL\_Large\_EK & 35.53 & 36.90 & $-1.37$ \\
19\phantom{0}(-) & RESDSQL\_Base\_EK & 29.14 & 31.16 & $-2.02$ \\
20\phantom{0}(-) & GPT-4-turbo\_predict\_dev & 25.68 & 22.75 & $+2.93$ \\
\bottomrule
\end{tabular}
}
\end{table}

\subsection{FP and TN Error Analysis}
\label{sec:appendix_error_cat}

\begin{figure*}[!ht]
    \centering
    \begin{subfigure}{0.49\textwidth}
        \centering
        \includegraphics[width=\textwidth]{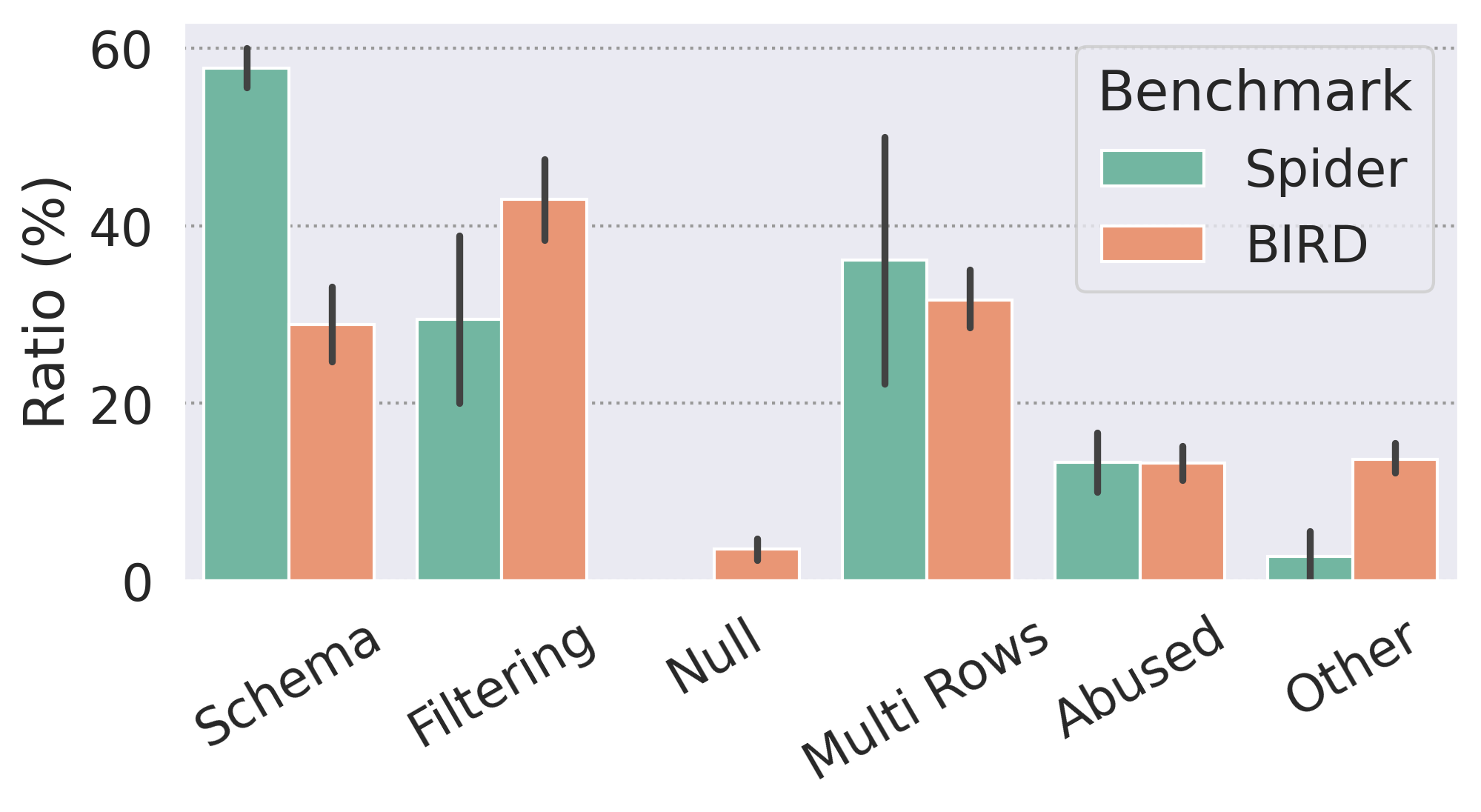}
        \caption{False positives}
        \label{fig:cat_top10_fp}
    \end{subfigure}
    \begin{subfigure}{0.49\textwidth}
        \centering
        \includegraphics[width=\textwidth]{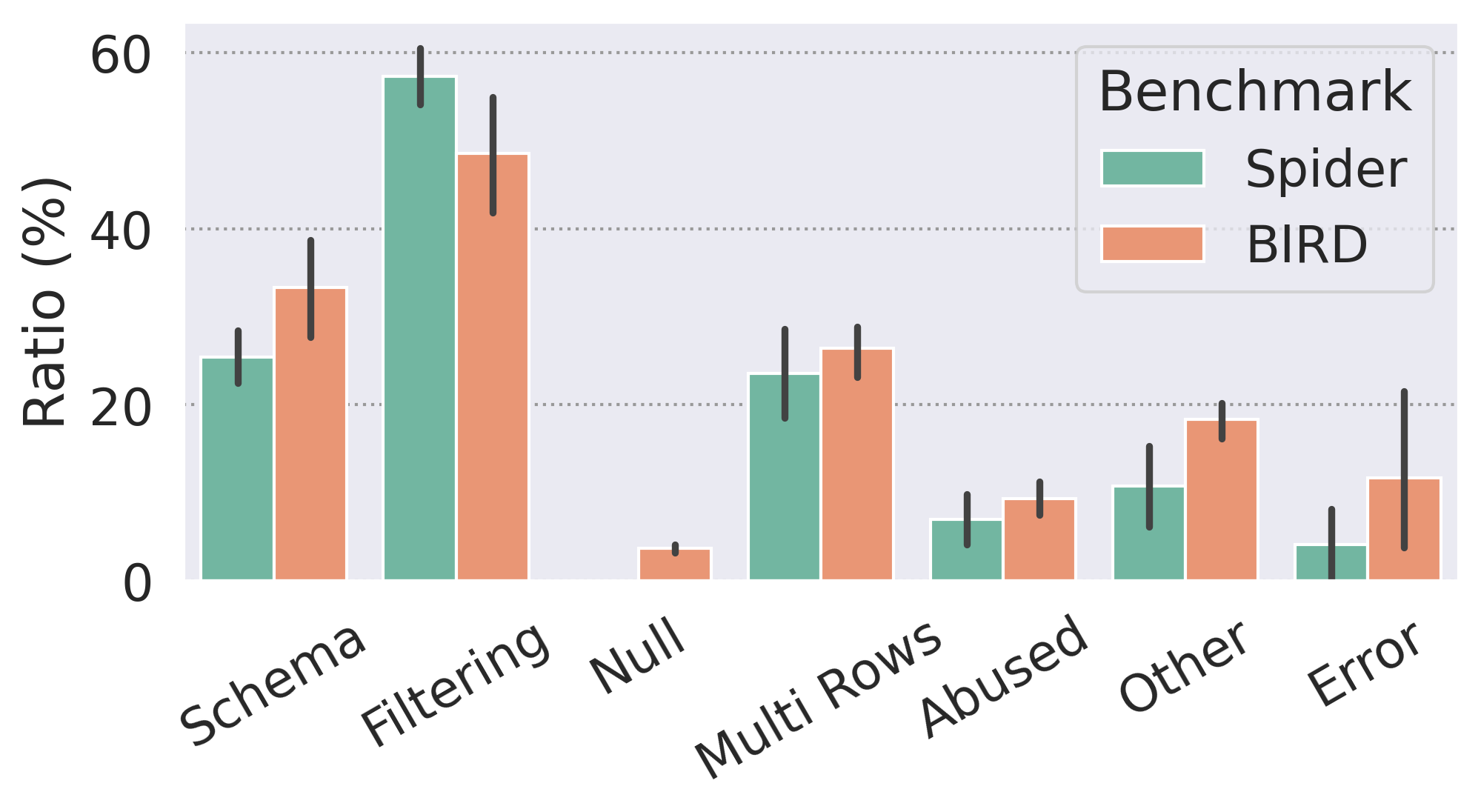}
        \caption{True negatives}
        \label{fig:cat_top10_tn}
    \end{subfigure}
    \caption{Average error categorization result of top 10 Spider-dev and BIRD-dev models using \ours. The gray line denotes the standard deviation. Schema denotes incorrect schema alignment, Filtering denotes incorrect filtering conditions, Null denotes incorrect handling of nullable columns, Multi Rows denotes incorrect accounting of multiple rows, Abused denotes abused clauses, and Error denotes SQL compile error.}
    \label{fig:cat_top10}
\end{figure*}

We generate a comprehensive evaluation report after obtaining judgments for all generated queries in a dataset.
This report includes an overall accuracy score, a breakdown of error types, and detailed explanations for each incorrect query. 
The LLM is prompted to categorize the error case based on its judgment and other contextual information.
Judgments are categorized according to the criteria described in the previous subsection, and complex errors can belong to more than one category.
Researchers can utilize this evaluation report to identify systematic errors in their text-to-SQL models and guide future improvements. 
Fig.~\ref{fig:fig_system_prompt_cat_fp} describes the system prompt to categorize false positives, and Fig.~\ref{fig:fig_system_prompt_cat_fn} describes the system prompt to categorize false negatives.
Fig.~\ref{fig:cat_top10} demonstrates the error categorization results of the top 10 BIRD-dev and Spider-dev models.

% To validate the categorization results, we sampled 100 false negatives and 100 false positives from categorized results of SuperSQL and CHESS-GPT-4o-mini using GPT-4o, and three human experts annotated the correctness of the categorized results. The overall agreement between GPT-4o and experts' consensus yielded Cohen's kappa of 67.85, indicating substantial agreement. Inter-annotator agreement yielded 59.71, indicating moderate agreement. 


\section{EX Error Case Study}
\label{sec:appendix_errors}
We sampled the representative examples from our human study. Table~\ref{tab:qual_incorrect_schema}, \ref{tab:qual_incorrect_condition}, \ref{tab:qual_abused_clause} and \ref{tab:qual_null_aggre} demonstrate the false positive examples. Table~\ref{tab:qual_extra_col}, \ref{tab:qual_column_order}, \ref{tab:qual_different_output_repr}, \ref{tab:qual_different_output_structure} demonstrate the false negative examples. Table~\ref{tab:qual_incorrect_gt} demonstrates the example, including different column ordering and incorrect ground truth.

\begin{table*}
\centering
\caption{Equivalent set \#3 (BIRD \#9): An incorrect schema linking case demonstrating a generated query that mistakenly uses 'schools.FundingType' instead of 'T2.Charter Funding Type'. The human judgment accurately identifies this schema linking error, further corroborated by the detailed FLEX w/ GPT-4o analysis.}
\label{tab:qual_incorrect_schema}
\adjustbox{max width=\textwidth}{
\begin{tblr}{
  vline{2} = {1-5}{},
  hline{1,6} = {-}{0.08em},
  hline{2,4-5} = {-}{},
}
Question           & Among the schools with the average score in Math over 560 in the SAT test, how many schools are directly charter-funded?                                                                                                                                                                                                                                                                                                                                                                                                                                                                                                                                                                                                                                                                                                                                                                                                                                                                                                                                                                                                                                                                                                                                                                                                                                                                                                                                                                                                                                                                                                                                                        \\
Ground Truth Query & {SELECT COUNT(T2.̀School Codè) FROM satscores AS T1 INNER JOIN frpm AS T2 ON T1.cds = T2.CDSCode\\WHERE T1.AvgScrMath  560 AND T2.̀Charter Funding Typè = \`{}Directly funded'\\\textgreater{} 2}                                                                                                                                                                                                                                                                                                                                                                                                                                                                                                                                                                                                                                                                                                                                                                                                                                                                                                                                                                                                                                                                                                                                                                                                                                                                                                                                                                                                                                                                             \\
Generated Query    & {SELECT COUNT(*) FROM satscores INNER JOIN schools ON satscores.cds = schools.CDSCode\\WHERE satscores.AvgScrMath  560    AND schools.FundingType = \`{}Directly funded';\\\textgreater{} 2}                                                                                                                                                                                                                                                                                                                                                                                                                                                                                                                                                                                                                                                                                                                                                                                                                                                                                                                                                                                                                                                                                                                                                                                                                                                                                                                                                                                                                                                                                      \\
Human Judgment     & {Incorrect Schema Linking: The generated query uses incorrect columns schools.̀funding typè\\The generated query is incorrect}                                                                                                                                                                                                                                                                                                                                                                                                                                                                                                                                                                                                                                                                                                                                                                                                                                                                                                                                                                                                                                                                                                                                                                                                                                                                                                                                                                                                                                                                                                                                                 \\
{\ours\ \textit{w/ GPT-4o}\\\textbf{\textcolor{blue}{(Correct)}}}     & {The Prediction Query and the Ground Truth Query both aim to count the number of schools that meet certain criteria. \\However, they differ in the tables and columns they use to achieve this.\\1. Tables and Joins:\\~ ~ - The Prediction Query uses the satscores and schools tables, joining them on satscores.cds = schools.CDSCode.\\~ ~ -~The Ground Truth Query uses the satscores and frpm tables, joining them on T1.cds = T2.CDSCode.\\2.~ Conditions:\\~ ~ -~Both queries filter for schools with an average Math score greater than 560.\\~ ~ -~The Prediction Query checks if schools.FundingType is 'Directly funded'.\\~ ~ -~The Ground Truth Query checks if frpm.Charter Funding Type is 'Directly funded'.\\3. Schema Considerations:\\~ ~ -~The hint indicates that schools.FundingType and frpm.Charter Funding Type are not equivalent, \\~ ~ ~ ~which means they may not represent the same data or criteria.\\~ ~ -~The Ground Truth Query uses frpm.Charter Funding Type, which is the correct column \\~ ~ ~ ~to check for charter funding type according to the question.\\4. Logical Correctness:\\~ ~ -~ The Prediction Query uses a different column (schools.FundingType) to determine if a school is directly charter-funded, \\~ ~ ~ ~ which is not equivalent to frpm.Charter Funding Type. \\~ ~ ~ ~ This could lead to incorrect results if the data in these columns differ.\\\\Given these points, the Prediction Query does not logically answer the question as it uses an incorrect column \\to determine the charter funding type. Therefore, the Prediction Query is incorrect.\\\`{}\`{}\`{}json\\\{"correct": false\}\\\`{}\`{}\`{}} 
\end{tblr}
}
\end{table*}

\begin{table*}
\centering
\caption{Equivalent set \#28 (BIRD \#68):  An incorrect condition filtering case demonstrating a generated query that erroneously adds 'StatusType = 'Closed'' and uses 'SOC = '11'' as a string instead of an integer. The human judgment correctly identifies the missing 'StatusType = 'Closed'' condition, while FLEX w/ GPT-4o provides a more comprehensive analysis, highlighting both the SOC value type discrepancy and the missing StatusType condition. This example emphasizes the critical nature of precise condition filtering in SQL queries.}
\label{tab:qual_incorrect_condition}
\adjustbox{max width=\textwidth}{
\begin{tblr}{
  vline{2} = {-}{},
  hline{1,6} = {-}{0.08em},
  hline{2,4-5} = {-}{},
}
Question           & {Which county reported the most number of school closure in the 1980s with school wonership code\\belonging to Youth Authority Facilities (CEA)?}                                                                                                                                                                                                                                                                                                                                                                                                                                                                                                                                                                                                                                                                                         \\
Ground Truth Query & {SELECT schools.County FROM schools WHERE schools.SOC = \`{}11'\\~AND strftime(\`{}\%Y', schools.ClosedDate) BETWEEN \`{}1980' AND '1989'\\GROUP BY schools.County ORDER BY COUNT(schools.County) DESC LIMIT 1;\\\textgreater{} Sacramento}                                                                                                                                                                                                                                                                                                                                                                                                                                                                                                                                                                                                   \\
Generated Query    & {SELECT County FROM schools WHERE strftime(\`{}\%Y', ClosedDate) BETWEEN \`{}1980' AND \`{}1989'    \\~AND StatusType = \`{}Closed'    AND SOC = 11\\GROUP BY County ORDER BY COUNT(School) DESC LIMIT 1~~\\\textgreater{} Sacramento}                                                                                                                                                                                                                                                                                                                                                                                                                                                                                                                                                                                                        \\
Human Judgment     & {Incorrect Condition Filtering: The generated query misses a filtering condition StatusType = \`{}Closed'\\The generated query is incorrect}                                                                                                                                                                                                                                                                                                                                                                                                                                                                                                                                                                                                                                                                                              \\
{\ours\ \textit{w/ GPT-4o}\\\textbf{\textcolor{blue}{(Correct)}}}     & {\`{}\`{}\`{}json\\\{"correct": false\}\\\`{}\`{}\`{}\\The Prediction Query is incorrect for the following reasons:\\1. SOC Value Type: The Prediction Query uses SOC = '11' with '11' as a string, while the Ground Truth Query \\uses SOC = 11 as an integer. This could lead to different results if the SOC column is of integer type.\\2. StatusType Condition: The Prediction Query does not include the condition StatusType = 'Closed', \\which is present in the Ground Truth Query. This condition is crucial to ensure that only closed schools are considered, \\which directly affects the result.\\These differences mean that the Prediction Query could potentially include incorrect data, leading to different results \\if the data set includes schools that are not closed or if the SOC column is of integer type.} 
\end{tblr}
}
\end{table*}

\begin{table*}
\centering
\caption{Equivalent set \#66 (BIRD \#45): This case showcases a generated query that unnecessarily includes a GROUP BY clause. Human judgment correctly identifies this issue, noting that it can yield incorrect results if there are multiple schools with the same name. FLEX w/ GPT-4o incorrectly assesses both queries as equivalent, overlooking the potential problems caused by the superfluous GROUP BY clause. This example highlights the importance of careful query structure analysis and the challenges in detecting subtle logical errors in SQL query generation and evaluation.}
\label{tab:qual_abused_clause}
\adjustbox{max width=\textwidth}{
\begin{tblr}{
  vline{2} = {-}{},
  hline{1,6} = {-}{0.08em},
  hline{2,4-5} = {-}{},
}
Question                                       & {What is the average writing score of each of the schools managed by Ricci Ulrich? \\List the schools and the corresponding average writing scores.}                                                                                                                                                                                                                                                                                                                                                                                                                                                                                                                                                                                                                                                                                                                                                                                                                                                                                                                                                                                                                                                                                                                                                                                                                                                                                                                                                                                                                                                                                                                                                                                                                                                                                                                                                                                                             \\
Ground Truth Query                             & {SELECT T2.School, T1.AvgScrWrite FROM satscores AS T1\\INNER JOIN schools AS T2 ON T1.cds = T2.CDSCode\\WHERE T2.AdmFName1 = \`{}Ricci' AND T2.AdmLName1 = \`{}Ulrich'\\\textgreater{} \textbar{} Buchanan High \textbar{} 507.0 \textbar{}}                                                                                                                                                                                                                                                                                                                                                                                                                                                                                                                                                                                                                                                                                                                                                                                                                                                                                                                                                                                                                                                                                                                                                                                                                                                                                                                                                                                                                                                                                                                                                                                                                                                                                                                     \\
Generated Query                                & {SELECT T1.sname, AVG(T1.avgscrwrite) FROM satscores AS T1\\INNER JOIN schools AS T2 ON T1.cds = T2.cdscode\\WHERE T2.admfname1 = \`{}Ricci' AND T2.admlname1 = \`{}Ulrich' GROUP BY T1.sname\\\textgreater{} \textbar{} Buchanan High \textbar{} 507.0 \textbar{}}                                                                                                                                                                                                                                                                                                                                                                                                                                                                                                                                                                                                                                                                                                                                                                                                                                                                                                                                                                                                                                                                                                                                                                                                                                                                                                                                                                                                                                                                                                                                                                                                                                                                                               \\
Human Judgment                                 & {The generated query misses a GROUP BY clause. It can yield different results\\If there are several schools with the same name.\\The generated query is incorrect.}                                                                                                                                                                                                                                                                                                                                                                                                                                                                                                                                                                                                                                                                                                                                                                                                                                                                                                                                                                                                                                                                                                                                                                                                                                                                                                                                                                                                                                                                                                                                                                                                                                                                                                                                                                                              \\
{\ours\ \textit{w/ GPT-4o}\\\textcolor{red}{(Incorrect)}} & {To evaluate the correctness of the Prediction Query, let's analyze both the Prediction Query \\and the Ground Truth Query in the context of the provided schema and question.\\Objective: The question asks for the average writing score of each school managed by Ricci Ulrich, \\listing the schools and their corresponding average writing scores.\\Schema Understanding:\\- satscores table contains cds and avgscrwrite.\\- schools table contains cdscode, sname, admfname1, and admlname1.\\Ground Truth Query Analysis:\\- It selects School and AvgScrWrite from the satscores and schools tables.\\- It joins these tables on T1.cds = T2.CDSCode.\\- It filters for schools managed by Ricci Ulrich using T2.AdmFName1 = 'Ricci' AND T2.AdmLName1 = 'Ulrich'.\\Prediction Query Analysis:\\- It selects T1.sname and the average of T1.avgscrwrite.\\- It joins the satscores and schools tables on T1.cds = T2.cdscode.\\- It filters for schools managed by Ricci Ulrich using T2.admfname1 = 'Ricci' and T2.admlname1 = 'Ulrich'.\\- It groups the results by T1.sname to calculate the average writing score for each school.\\Comparison:\\- The Prediction Query correctly joins the satscores and schools tables and filters for the correct administrator.\\- The use of AVG(T1.avgscrwrite) and GROUP BY T1.sname in the Prediction Query is appropriate \\to calculate the average writing score for each school.\\- The Ground Truth Query does not perform an aggregation, \\which suggests it assumes precomputed average scores are available in T1.AvgScrWrite.\\Result Consistency:\\- Both queries produce the same results, indicating that the Prediction Query correctly calculates the average writing scores.\\Given the analysis, the Prediction Query logically answers the question and produces the correct results. \\Therefore, the Prediction Query is correct.\\\`{}\`{}\`{}json\\\{"correct": true\}\\\`{}\`{}\`{}} 
\end{tblr}

}
\end{table*}

\begin{table*}
\centering
\caption{Equivalent set \#73 (BIRD \#64): A case highlighting the importance of handling nullable columns in SQL queries. The human correctly identifies that COUNT(*) and COUNT(charternum) treat null values differently, potentially affecting the result if charternum is nullable. FLEX w/ GPT-4o incorrectly deems the queries equivalent, overlooking this crucial distinction. This example reveals a limitation in the AI system's query analysis, emphasizing the need for careful consideration of null value handling in SQL evaluation.}
\label{tab:qual_null_aggre}
\adjustbox{max width=\textwidth}{
\begin{tblr}{
  vline{2} = {-}{},
  hline{1,6} = {-}{0.08em},
  hline{2,4-5} = {-}{},
}
Question                                       & {What is the total number of non-chartered schools in the county of Los Angeles \\with a percent (\%) of eligible free meals for grades 1 through 12 that is less than 0.18\%?}                                                                                                                                                                                                                                                                                                                                                                                                                                                                                                                                                                                                                                                                                                                                                                                                                                                                             \\
Ground Truth Query                             & {SELECT COUNT(charternum) FROM schools WHERE mailcity = \`{}Hickman'    AND charternum = \`{}00D4'\\\textgreater{} 3}                                                                                                                                                                                                                                                                                                                                                                                                                                                                                                                                                                                                                                                                                                                                                                                                                                                                                                                                       \\
Generated Query                                & {SELECT COUNT(*) FROM schools WHERE CharterNum = \`{}00D4'    AND MailCity = \`{}Hickman'\\\textgreater{} 3}                                                                                                                                                                                                                                                                                                                                                                                                                                                                                                                                                                                                                                                                                                                                                                                                                                                                                                                                                 \\
Human Judgment                                 & {Missing handling of the nullable column: charternum is the nullable column in the schema\\COUNT(*) counts the null values while COUNT(charternum) does not.\\The generated query is incorrect.}                                                                                                                                                                                                                                                                                                                                                                                                                                                                                                                                                                                                                                                                                                                                                                                                                                                            \\
{\ours\ \textit{w/ GPT-4o}\\\textcolor{red}{(Incorrect)}} & {The Prediction Query and the Ground Truth Query are essentially the same in terms of logic and execution. \\Both queries aim to count the number of schools in the 'schools' table where the 'MailCity' is 'Hickman' and the 'CharterNum' is '00D4'.\\Key points to consider:\\- Both queries use the same filtering conditions: mailcity = 'Hickman' and charternum = '00D4'.\\- The difference in column case sensitivity (charternum vs CharterNum and mailcity vs MailCity) is not an issue in SQLite, \\as it is case-insensitive with respect to column names.\\- The use of COUNT(charternum) in the Prediction Query versus COUNT(*) in the Ground Truth Query does not affect the result \\as long as charternum is not nullable. If charternum were nullable, it could potentially affect the count, but the question implies that \\it is not an issue here since both queries produce the same result.\\Given these considerations, the Prediction Query correctly answers the question.\\\`{}\`{}\`{}json\\\{"correct": true\}\\\`{}\`{}\`{}} 
\end{tblr}
}
\end{table*}

\begin{table*}
\centering
\caption{Not equivalent set \#90 (BIRD \#63): A case showcases FLEX's ability to evaluate SQL queries with different column selections correctly. The Generated Query includes extra columns for additional administrator names (admname2, admname3) that are not present in the Ground Truth Query. Despite this difference, FLEX accurately judges that both queries are logically equivalent in answering the question about administrators in chartered schools with charter number '0002'. This demonstrates FLEX's sophisticated understanding that the presence of additional, non-conflicting information does not alter the fundamental correctness of a query in meeting the specified requirements.}
\label{tab:qual_extra_col}
\adjustbox{max width=\textwidth}{
\begin{tblr}{
  vline{2} = {-}{},
  hline{1,6} = {-}{0.08em},
  hline{2,4-5} = {-}{},
}
Question                                               & {In chartered schools with charter number 00D2, what are the names of all the administrators? \\Include the name of the school and the city to which it belongs}                                                                                                                                                                                                                                                                                                                                                                                                                                                                                                                                                                                                                                                                                                                                                      \\
Ground Truth Query                                     & {SELECT AdmFName1, AdmLName1, School, City FROM schools\\WHERE Charter = 1 AND CharterNum = \`{}00D2'}                                                                                                                                                                                                                                                                                                                                                                                                                                                                                                                                                                                                                                                                                                                                                                                                                 \\
Generated Query                                        & {SELECT school, city, admfname1, admlname1, admfname2, admlname2, admfname3, admlname3 FROM schools\\WHERE charter = 1 AND charternum = \`{}00D2'}                                                                                                                                                                                                                                                                                                                                                                                                                                                                                                                                                                                                                                                                                                                                                                    \\
Human Judgment                                         & {Extra Column:~admfname2, admlname2, admfname3, admlname3, but the other logic is correct.\\The generated query is correct.}                                                                                                                                                                                                                                                                                                                                                                                                                                                                                                                                                                                                                                                                                                                                                                                          \\
{\ours\ \textit{w/ GPT-4o}\\\textbf{\textcolor{blue}{(Correct)}}} & {The task is to find the names of all administrators in chartered schools with charter number '00D2', \\including the name of the school and the city to which it belongs.\\Let's compare the Prediction Query and the Ground Truth Query:\\Filtering Condition:\\- Both queries correctly filter the schools where Charter = 1 and CharterNum = '00D2'. \\- This part of the query is consistent with the requirement to focus on chartered schools with the specified charter number.
\\Selected Columns:\\- The Ground Truth Query selects AdmFName1, AdmLName1, School, and City. \\- 
 This directly answers the question by providing the first administrator's first and last name, along with the school and city.\\- The Prediction Query selects school, city, admfname1, admlname1, admfname2, admlname2, admfname3, admlname3.\\- This includes additional columns for the second and third administrators' names, which are not required by the question.\\However, this does not affect the logical correctness of the query as it still provides the necessary information.\\Output Structure:\\- The Ground Truth Result has four columns,\\ while the Prediction Result has eight columns due to the inclusion of additional administrator name fields.\\- The extra columns in the Prediction Result do not impact the ability to answer the question, as the required information is still present.\\Result Comparison:\\- The results from both queries contain the same schools and cities, and the first administrator's names match.\\-  The additional columns in the Prediction Result are filled with None for the second and third administrators, \\which does not affect the correctness of the answer to the question.\\- In conclusion, the Prediction Query logically answers the question by providing the necessary information\\ about the first administrator's name, school, and city, even though it includes additional columns.\\- The extra columns do not introduce any errors or inconsistencies in the context of the question.\\\`{}\`{}\`{}json\\\{"correct": true\}\\\`{}\`{}\`{}} 
\end{tblr}
}
\end{table*}

\begin{table*}
\centering
\caption{Not equivalent set \#17 (BIRD \#37): A case where FLEX correctly judges two SQL queries as logically equivalent despite a difference in column order. The Generated Query selects columns as T2.Street, T2.City, T2.State, T2.Zip, while the Ground Truth Query selects them as Street, City, State, Zip. Despite this variation in the SELECT clause, both queries aim to find the complete address of the school with the lowest excellence rate. FLEX accurately determines that the column order difference does not affect the logical correctness of the query in answering the question, demonstrating its ability to evaluate query semantics beyond superficial syntax.}
\label{tab:qual_column_order}
\adjustbox{max width=\textwidth}{

\begin{tblr}{
  vline{2} = {-}{},
  hline{1-2,4-5} = {-}{},
  hline{1,6} = {-}{0.08em},
}
Question                                               & What is the complete address of the school with the lowest excellence rate? Indicate the Street, City, Zip and State.                                                                                                                                                                                                                                                                                                                                                                                                                                                                                                                                                                                                                                                                                                                                                                                                                                                                                                                                                                                                                                                                                                                                                                                                                                                                                                                                                                                                                                                                                                                                                                                                                                                        \\
Ground Truth Query                                     & {SELECT schools.Street, schools.City, schools.Zip, schools.State FROM schools\\INNER JOIN satscores ON schools.CDSCode = satscores.cds\\WHERE satscores.NumTstTakr != 0\\ORDER BY (satscores.NumGE1500 * 1.0 / satscores.NumTstTakr) ASC LIMIT 1;\\\textgreater{} | 1111 Van Ness Avenue | Fresno | CA | 93721-2002 |}                                                                                                                                                                                                                                                                                                                                                                                                                                                                                                                                                                                                                                                                                                                                                                                                                                                                                                                                                                                                                                                                                                                                                                                                                                                                                                                                                                                                                                                                                                            \\
Generated Query                                        & {SELECT T2.Street, T2.City, T2.State, T2.Zip FROM satscores AS T1\\INNER JOIN schools AS T2 ON T1.cds = T2.CDSCode\\ORDER BY CAST(T1.NumGE1500 AS REAL) / T1.NumTstTakr ASC LIMIT 1\\\textgreater{} | 1111 Van Ness Avenue | Fresno | 93721-2002 | CA |}                                                                                                                                                                                                                                                                                                                                                                                                                                                                                                                                                                                                                                                                                                                                                                                                                                                                                                                                                                                                                                                                                                                                                                                                                                                                                                                                                                                                                                                                                                                                                                          \\
Human Judgment                                         & Only the column order is different, so the generated query is correct.                                                                                                                                                                                                                                                                                                                                                                                                                                                                                                                                                                                                                                                                                                                                                                                                                                                                                                                                                                                                                                                                                                                                                                                                                                                                                                                                                                                                                                                                                                                                                                                                                                                                                                       \\
{\ours\ \textit{w/ GPT-4o}\\\textbf{\textcolor{blue}{(Correct)}}} & {The task is to find the complete address of the school with the lowest excellence rate,\\ defined as NumGE1500 / NumTstTakr. The complete address should include the Street, City, State, and Zip code.\\

Let's compare the Prediction Query and the Ground Truth Query:\\

Join Condition:\\
- Both queries correctly join the schools and satscores tables using the CDSCode and cds columns, respectively.\\

Excellence Rate Calculation:\\
- Both queries calculate the excellence rate as NumGE1500 / NumTstTakr.\\ - The Ground Truth Query explicitly casts NumGE1500 as REAL to ensure floating-point division, \\while the Prediction Query multiplies NumGE1500 by 1.0 to achieve the same effect. - \\Both methods are valid for ensuring floating-point division in SQLite.\\

Filtering:\\
- The Prediction Query includes a condition satscores.NumTstTakr != 0 to avoid division by zero,\\which is a good practice. \\- The Ground Truth Query does not explicitly include this condition, \\but it is implied that NumTstTakr should not be zero for a valid excellence rate calculation.\\

Ordering and Limiting:\\
- Both queries order the results by the calculated excellence rate in ascending order and limit the result to the first entry, \\which corresponds to the school with the lowest excellence rate.\\

Selected Columns:\\
- The Prediction Query selects Street, City, Zip, State, while the Ground Truth Query selects Street, City, State, Zip. \\The order of columns in the SELECT clause does not affect the correctness of the query in terms of answering the question.\\

Result Discrepancy:\\
- The Prediction Result and Ground Truth Result differ in the order of the State and Zip columns.\\ However, this does not affect the logical correctness of the query in terms of providing the complete address.\\
- In conclusion, the Prediction Query logically answers the question by providing the complete address of the school with the lowest excellence rate. \\ - The difference in column order in the result does not impact the correctness of the query.\\\`{}\`{}\`{}json\\\{"correct": true\}\\\`{}\`{}\`{}} 
\end{tblr}
}
\end{table*}

\begin{table*}
\centering
\caption{Not equivalent set \#16 (BIRD \#32): A different output value case demonstrates the accuracy of FLEX judgment in SQL query evaluation. While the Generated Query used the ROUND function to present results as percentages, the Ground Truth Query expressed them as decimal values. Despite this difference in output format, FLEX correctly determined that both queries produce logically equivalent results. This showcases FLEX's ability to understand and evaluate query semantics and functionality beyond simple text comparison.}
\label{tab:qual_different_output_repr}
\adjustbox{max width=\textwidth}{
\begin{tblr}{
  vline{2} = {-}{},
  hline{1,6} = {-}{0.08em},
  hline{2,4-5} = {-}{},
}
Question                                               & {What is the eligible free or reduced price meal rate for the top 5 schools in grades 1-12 \\with the highest free or reduced price meal count of the schools with the ownership code 66?}                                                                                                                                                                                                                                                                                                                                                                                                                                                                                                                                                                                                                                                                                                                                                                                                                                                                                                                                                                                     \\
Ground Truth Query                                     & {SELECT CAST(T1.̀FRPM Count (K-12)̀ AS REAL) / T1.̀Enrollment (K-12)̀ FROM frpm AS T1\\INNER JOIN schools AS T2 ON T1.CDSCode = T2.CDSCode WHERE T2.SOC = 66\\ORDER BY T1.̀FRPM Count (K-12)̀ DESC LIMIT 5\\\textgreater{} \textbar{} 0.9993~\textbar{} 0.9179~\textbar{} 0.9142 \textbar{} 0.8965 \textbar{} 0.8958 \textbar{}}                                                                                                                                                                                                                                                                                                                                                                                                                                                                                                                                                                                                                                                                                                                                                                                                                                                 \\
Generated Query                                        & {SELECT ROUND((frpm.̀FRPM Count (K-12)̀ / frpm.̀Enrollment (K-12)̀) * 100, 2) AS eligible\_rate FROM frpm\\INNER JOIN schools ON frpm.CDSCode = schools.CDSCode WHERE schools.SOC = \`{}66'\\ORDER BY frpm.̀FRPM Count (K-12)̀ DESC LIMIT 5\\\textgreater{} \textbar{} 99.93 \textbar{} 91.79 \textbar{} 91.42~\textbar{} 89.65 \textbar{}~89.58 \textbar{}}                                                                                                                                                                                                                                                                                                                                                                                                                                                                                                                                                                                                                                                                                                                                                                                                                   \\
Human Judgment                                         & {Different output value representation: The generated query multiples 100 to the result for readability.\\Other logic is correct; the generated query is correct}                                                                                                                                                                                                                                                                                                                                                                                                                                                                                                                                                                                                                                                                                                                                                                                                                                                                                                                                                                                                              \\
{\ours\ \textit{w/ GPT-4o}\\\textbf{\textcolor{blue}{(Correct)}}} & {The Prediction Query and the Ground Truth Query both aim to calculate the eligible free or reduced price meal rate for the top 5 schools \\ with the highest free or reduced price meal count,\\ specifically for schools with the ownership code 66. However, there are some differences in their execution and results:\\

Calculation of Eligible Rate:\\
- The Ground Truth Query calculates the eligible rate as a floating-point number by dividing FRPM Count (K-12) by Enrollment (K-12).\\
- The Prediction Query also performs this division but multiplies the result by 100 and rounds it to two decimal places,\\ effectively converting the rate into a percentage.\\

Result Representation:\\
- The Ground Truth Result presents the eligible rates as decimal values (e.g., 0.8964987714987716).\\
- The Prediction Result presents the eligible rates as percentages (e.g., 89.65).\\

Logical Correctness:\\
- Both queries logically answer the question by selecting the top 5 schools based on the FRPM Count (K-12) and \\calculating the eligible rate for these schools.\\
- The difference in representation (percentage vs. decimal) does not affect the logical correctness of the Prediction Query.\\ 
- The conversion to percentage is a valid transformation of the rate.\\

SQLite3 Compatibility:\\
- Both queries are compatible with SQLite3, and the operations performed are valid within this context.\\
- In conclusion, the Prediction Query is correct in terms of logic and functionality.\\ 
- The difference in result representation (percentage vs. decimal) is acceptable and does not impact the query's ability to answer the question.} \\
\end{tblr}
}
\end{table*}

\begin{table*}
\centering
\caption{Not equivalent set \#26 (BIRD \#53): This case illustrates FLEX's capability to correctly evaluate queries with different output structures but equivalent logical intent. The Generated Query uses SUM to aggregate the total number of test takers in Fresno schools, while the Ground Truth Query lists individual values for each school. Despite this structural difference, FLEX accurately judges that the Generated Query correctly answers the question, "How many test takers are at the schools whose mailing city address is in Fresno?". This demonstrates FLEX's understanding that aggregation can be an appropriate method to answer quantitative questions, even when the reference query provides more granular data.}
\label{tab:qual_different_output_structure}
\adjustbox{max width=\textwidth}{
\begin{tblr}{
  vline{2} = {-}{},
  hline{1,6} = {-}{0.08em},
  hline{2,4-5} = {-}{},
}
Question                                               & How many test takers are there at the school/s whose mailing city address is in Fresno?                                                                                                                                                                                                                                                                                                                                                                                                                                                                                                                                                                                                                                                                                                                                                                                                                                                                                                                                                                                                                                                                                                                                                                                                                                                                             \\
Ground Truth Query                                     & {SELECT T1.NumTstTakr FROM satscores AS T1 \\INNER JOIN schools AS T2 ON T1.cds = T2.CDSCode WHERE T2.MailCity = \`{}Fresno'\\\textgreater{} \textbar{} 5 \textbar{} 8 \textbar{} 368 \textbar{} ....~ (Sum is 6070)}                                                                                                                                                                                                                                                                                                                                                                                                                                                                                                                                                                                                                                                                                                                                                                                                                                                                                                                                                                                                                                                                                                                                               \\
Generated Query                                        & {SELECT SUM(satscores.NumTstTakr) FROM schools\\INNER JOIN satscores ON schools.CDSCode = satscores.cds WHERE schools.MailCity = \`{}Fresno'; 6070\\\textgreater{} 6070}                                                                                                                                                                                                                                                                                                                                                                                                                                                                                                                                                                                                                                                                                                                                                                                                                                                                                                                                                                                                                                                                                                                                                                                              \\
Human Judgment                                         & {Different output structure: Two queries have the same JOIN and WHERE structure, \\but the generated query adds all values while the ground truth does not.\\The question does not specify it, so the generated query is correct.}                                                                                                                                                                                                                                                                                                                                                                                                                                                                                                                                                                                                                                                                                                                                                                                                                                                                                                                                                                                                                                                                                                                                  \\
{\ours\ \textit{w/ GPT-4o}\\\textbf{\textcolor{blue}{(Correct)}}} & {The task is to determine the total number of test takers at schools whose mailing city address is in Fresno.\\

Let's compare the two queries:\\

Prediction Query:\\
- This query calculates the sum of the NumTstTakr column from the satscores table for all schools in Fresno. \\
- It uses an INNER JOIN to match records from the schools and satscores tables based on the CDSCode and cds columns, respectively. \\
- The result is a single aggregated value representing the total number of test takers in Fresno.\\

Ground Truth Query:\\
- This query retrieves individual NumTstTakr values for each school in Fresno without aggregation. \\ - It also uses an INNER JOIN to match records from the schools and satscores tables.\\
- The result is a list of NumTstTakr values for each school in Fresno.\\

Analysis: \\
- The Prediction Query provides the total number of test takers by summing up the NumTstTakr values,\\ which directly answers the question of "how many test takers are there" in total.\\
- The Ground Truth Query lists the number of test takers for each school individually, \\which does not directly answer the question of the total number of test takers.\\

Conclusion:\\
- The Prediction Query correctly answers the question by providing the total number of test takers in Fresno, \\while the Ground Truth Query provides a breakdown by school. \\ The difference in results is due to the aggregation in the Prediction Query, \\ which is appropriate for the question asked.\\\`{}\`{}\`{}json\\\{"correct": true\}\\\`{}\`{}\`{}} 
\end{tblr}

}
\end{table*}

\begin{table*}
\centering
\caption{Not equivalent set \#98 (BIRD \#80): A case highlighting the Ground Truth Query's incorrectness due to overlooking multiple rows satisfying the condition. The Ground Truth Query fails to account for potential ties in maximum latitude by using LIMIT 1. Despite this flaw, FLEX w/ GPT-4o incorrectly judges it as correct while correctly identifying the Generated Query's handling of ties. This reveals a limitation in the AI system's ability to recognize errors in Ground Truth Queries, particularly when dealing with maximum values and potential multiple matching rows in SQL evaluation..}
\label{tab:qual_incorrect_gt}
\adjustbox{max width=\textwidth}{
\begin{tblr}{
  vline{2} = {-}{},
  hline{1,6} = {-}{0.08em},
  hline{1-2,4-5} = {-}{},
}
Question                                                & {What is the school type of the school with the highest latitude? \\Indicate the name of the school as well as the latitude coordinates.}                                                                                                                                                                                                                                                                                                                                                                                                                                                                                                                                                                                                                                                                                                                                                                                                                                                                                                                                                                                                                                                                                                                                                                                                                                                                                                                                            \\
Ground Truth Query                                      & {SELECT T1.̀School Typè, T1.̀School Namè, T2.Latitude FROM frpm AS T1\\INNER JOIN schools AS T2 ON T1.CDSCode = T2.CDSCode\\ORDER BY T2.Latitude DESC LIMIT 1\\\textgreater{} \textbar{} High Schools (Public) \textbar{} Leggett Valley High \textbar{} 44.219305 \textbar{}}                                                                                                                                                                                                                                                                                                                                                                                                                                                                                                                                                                                                                                                                                                                                                                                                                                                                                                                                                                                                                                                                                                                                                                                                       \\
Generated Query                                         & {SELECT schools.school, schools.latitude, frpm.̀school typè FROM schools\\JOIN frpm ON schools.cdscode = frpm.cdscode\\WHERE schools.latitude =        (SELECT MAX(latitude)         FROM schools)\\ \textbar{} Leggett Valley High \textbar{} 44.219305 \textbar{} High Schools (Public) \textbar{}\\~ ~\textbar{} Leggett Valley Elementary \textbar{} 44.219305 \textbar{} Elementary Schools (Public) \textbar{}~~}                                                                                                                                                                                                                                                                                                                                                                                                                                                                                                                                                                                                                                                                                                                                                                                                                                                                                                                                                                                                                                                              \\
Human Judgment                                          & {The ground truth does not consider that multiple rows have the same highest latitude.\\Moreover, output column order is different, but the generated query is correct.}                                                                                                                                                                                                                                                                                                                                                                                                                                                                                                                                                                                                                                                                                                                                                                                                                                                                                                                                                                                                                                                                                                                                                                                                                                                                                                             \\
{\ours\ \textit{w/ GPT-4o}\\\textbf{\textcolor{red}{(Incorrect)}}} & {The task is to find the school type, name, and latitude of the school with the highest latitude. \\ Let's compare the Prediction Query and the Ground Truth Query to understand their differences and determine if the Prediction Query is correct.\\

Prediction Query Analysis\\
- This query joins the schools and frpm tables on the cdscode column.\\
- It selects the school name, latitude, and school type for schools with the maximum latitude.\\
- The subquery (SELECT MAX(latitude) FROM schools) correctly identifies the maximum latitude.\\
- The WHERE clause filters schools to only those with this maximum latitude.\\
- The result includes multiple rows if there are ties in the maximum latitude.\\

Ground Truth Query Analysis\\
- This query also joins the frpm and schools tables on the CDSCode column.\\
- It orders the results by latitude in descending order and limits the output to the top row, \\ - - - effectively selecting the school with the highest latitude.\\
- It selects the school type, school name, and latitude.\\

Comparison\\
- Result Set: The Prediction Query returns two rows because there are two schools with the same maximum latitude, \\while the Ground Truth Query returns only one row due to the LIMIT 1 clause.\\
- Columns: Both queries select the same logical information: school type, school name, and latitude, though the order of columns differs.\\
- Handling Ties: The Prediction Query correctly handles ties by returning all schools with the maximum latitude, \\which is logically consistent with the question's requirement to identify the school with the highest latitude.\\

Conclusion
- The Prediction Query logically answers the question by identifying all schools with the maximum latitude, \\which is a valid interpretation of the task. \\
- The Ground Truth Query limits the result to one school, which is another valid approach. \\
- The difference in the number of rows is due to the handling of ties, which is acceptable given the question's ambiguity.\\\`{}\`{}\`{}json\\\{"correct": true\}\\\`{}\`{}\`{}} 
\end{tblr}
}
\end{table*}

\section{Artifacts}
We confirmed that all models, datasets, and frameworks are allowed for research use. 

\section{AI Assitant}
We use \href{https://copilot.microsoft.com/}{Microsoft Copilot} as a coding assistant and \href{https://grammarly.com/}{Grammarly} and \href{https://www.writefull.com/}{Writefull} as a writing assistant.

\section{LLM Prompt}
\label{sec:appendix_prompt}
\subsection{FLEX}
\label{sec:appendix_prompt_flex}
We distinguished the prompt into system prompt and user prompt.
As described in Fig.~\ref{fig:fig_user_prompt}, the user prompt includes natural language questions, generated queries, ground truth queries, and execution results. The system prompt includes task instruction and criteria as described in Fig.~\ref{fig:fig_system_prompt_eq} and \ref{fig:fig_system_prompt_neq}. In the categorization, the user prompt includes natural language questions, generated queries, and ground truth queries. 
All system prompts instruct the LLM to output using Chain-of-Thought~\cite{wei2022chain} prompting followed by JSON formatted result to extract judgment from LLM output.

\subsection{Prometheus-2}
\label{sec:appendix_prompt_prometheus}

\prometheustwo~\cite{kim2024prometheus} requires instruction, response, reference answer, criteria, and score rubrics for evaluation. Fig.~\ref{fig:fig_prometheus_criteria_custom_eq} and \ref{fig:fig_prometheus_criteria_custom_neq} show optimal criteria for equivalent and not equivalent sets, respectively. We deemed scores of four and five as correct to achieve the highest level of agreement.

\onecolumn
\begin{figure}[!h]
    \centering
\begin{lstlisting}
The Prediction Result matches the Reference Result. However, this does not guarantee that the Prediction Query is correct. Carefully analyze the Prediction Query and evaluate its correctness considering the following criteria:

Correct Prediction Query
- If the Prediction Query missed some tables or columns, it is acceptable if the missing information does not affect the query's ability to answer the Question.
- missing DISTINCT in SELECT is always allowed because we do not consider duplicate rows.

Incorrect Prediction Query
- Does not logically answer the Question or contains significant errors.
- Produces different results due to incorrect filering or missing conditions, JOIN redundancy, or other fatal issues.
- Fails to handle null values, multiple rows, or other critical aspects of the Question.
- Does not consider nullable columns in aggregation functions (SUM, COUNT, AVG) and NULL value can lead to unexpected results.
For example, COUNT(*) in the prediction but COUNT(school) in the ground truth will produce different results if the column school does not have a NOT NULL constraint in the schema.
- Does not produce correct results when multiple rows satisfy the condition (e.g min/max, multiple transactions in a day).
- Abused clauses (LIMIT, GROUP BY) to limit the results when the user didn't request it.

SQLite3 Compatibility
- Both queries are SQLite3 compatible, meaning integer division is not automatically converted to float division, and logical operators, column/table names are case-insensitive.
- comparison between string and integer is allowed in SQLite3.

Analysis Guidelines
1. Compare the Prediction Query with the Ground Truth Query within the context of the provided schema and question.
2. Predict the query's logical correctness based on the criteria mentioned above.

Finally, score the Prediction Query as follows:
```json
{"correct": true or false}
```
\end{lstlisting}
    \caption{\ours\ system prompt to judge the generated query producing the same results as ground truth.}
    \label{fig:fig_system_prompt_eq}
\end{figure}

\begin{figure}[!h]
    \centering
\begin{lstlisting}
The Prediction Result differs from the Ground Truth Result. However, this does not necessarily mean that the Prediction Query is incorrect. Analyze the differences between the Prediction Query and Reference Query, considering the following:

Correct
- The Prediction Query logically answers the Question, even if the output structure differs from the Ground Truth Query.
- Do not consider column naming, column/row ordering.
- Some extra column or missing column in the output structure is acceptable if it does not affect the query's ability to answer the Question.
- Differences in the representation of values, such as formatting (percentile, YES/NO) or data types, are acceptable if they do not affect the query's logical correctness.
- Ambiguous questions may have multiple correct answers, so the Prediction Query may differ from the Ground Truth Query.
- Multiple rows are acceptable when the calculate the min/max

Incorrect
- The Prediction Query does not logically answer the Question or contains significant errors.
- The Prediction Query produces different results due to incorrect filering or missing conditions, JOIN redundancy, or other fatal issues.
- The Prediction Query fails to handle null values, multiple rows, or other critical aspects of the Question.
- The result of the Prediction Query is significantly different from the Ground Truth Query even its structure is similar.

SQLite3 Compatibility
- Both queries are SQLite3 compatible, meaning integer division is not automatically converted to float division, and logical operators are case-insensitive.
- If the table schema and description are different, follow the schema provided in the prompt.

Provide a detailed comparison of the Prediction Query and Ground Truth Query, focusing on the nature and significance of their differences. If the Prediction Query is incorrect, explain the specific errors and how they affect the query's ability to answer the Question.

Finally, score the Prediction Query as follows:
```json
{"correct": true/false}
```
\end{lstlisting}
    \caption{\ours\ system prompt to judge the generated query producing different results as ground truth.}
    \label{fig:fig_system_prompt_neq}
\end{figure}

\begin{figure}[!h]
    \centering
\begin{lstlisting}
You are an expert SQL evaluator responsible for assessing the correctness and quality of SQL queries. 
Analyze the judgment thoroughly and provide a categorized evaluation based on the following criteria:
Two queries have no compile-time errors, which means they are both valid SQL queries, integer division is not automatically converted to float division, and logical operators are case-insensitive.

### Criteria
1. Incorrect Schema Linking: Utilized tables and columns in the predicted query do not align with the question and the provided schema. Different columns are permissible if they are described as equivalent in the table description or if the question does not specify exact column names.
2. Incorrect Filtering Conditions: The prediction query incorrectly filters data based on the given conditions, ensuring the WHERE clause is used appropriately to match expected results.
3. Missing Handling of Nullable Column: Check if the query correctly handles nullable columns in aggregation functions (SUM, COUNT, AVG) or other operations, as improper handling can lead to unexpected results. Do not consider NULL values in arithmetic operations.
4. Missing Handling of Multiple Rows: Determine if the query correctly accounts for scenarios where multiple rows might satisfy the condition (e.g min/max, multiple transactions in a day), potentially leading to incorrect answers. Consider primary and foreign keys, as well as unique constraints, which ensure uniqueness.
5. Abused Clauses: Evaluate if SQL clauses like GROUP BY, HAVING, ORDER BY, and DISTINCT are used unnecessarily, which could produce incorrect results.
6. Other Fatal Logical Issues: Identify any additional logical problems not covered by the above criteria.


### Output Format
NOTE: If no issue is found, the explanation can be an empty string. Output only the JSON object containing your evaluation results. Make sure the JSON is properly formatted and valid, with all boolean values in lowercase (true/false) and all strings properly enclosed in double quotes.

```json
{
    "incorrect_schema_linking": {
        "issued": true/false,
        "explanation": "Your explanation here"
    },
    "incorrect_filtering_condition": {
        "issued": true/false,
        "explanation": "Your explanation here"
    },
    "missing_handling_of_nullable_column": {
        "issued": true/false,
        "explanation": "Your explanation here"
    },
    "missing_handling_of_multiple_rows": {
        "issued": true/false,
        "explanation": "Your explanation here"
    },
    "abused_clauses": {
        "issued": true/false,
        "explanation": "Your explanation here"
    },
    "other_fatal_issues": {
        "issued": true/false,
        "explanation": "Your explanation here"
    }
}
```
\end{lstlisting}
    \caption{\ours\ system prompt to categorize the error case of false positives.}
    \label{fig:fig_system_prompt_cat_fp}
\end{figure}

\begin{figure}[!h]
    \centering
\begin{lstlisting}
You are an expert SQL evaluator responsible for assessing the correctness and quality of SQL queries. 
Analyze the judgment thoroughly and provide a categorized evaluation based on the following criteria:
Two queries have no compile-time errors, which means they are both valid SQL queries, integer division is not automatically converted to float division, and logical operators are case-insensitive.

### Criteria
1. Different Output Structure: Column selection, ordering is different, but should not consider output column naming.
2. Different Output Value Representation: Differences in value representation, such as formatting or data types, should not be fatal unless they affect the logical correctness of the query.
3. Incorrect Ground Truth Query: The ground truth query is incorrect and does not logically answer the question.
4. Multiple Answers Available: The question is ambiguous and has multiple correct answers.
5. Other minor issues: Other minor issues that do not affect the logical correctness of the query.

### Output Format
NOTE: If no issue is found, the explanation can be an empty string. Output only the JSON object containing your evaluation results. Make sure the JSON is properly formatted and valid, with all boolean values in lowercase (true/false) and all strings properly enclosed in double quotes.

```json
{
    "different_output_structure": {
        "issued": true/false,
        "explanation": "Your explanation here"
    },
    "different_output_value_representation": {
        "issued": true/false,
        "explanation": "Your explanation here"
    },
    "incorrect_ground_truth_query": {
        "issued": true/false,
        "explanation": "Your explanation here"
    },
    "multiple_answers_available": {
        "issued": true/false,
        "explanation": "Your explanation here"
    },
    "other_minor_issues": {
        "issued": true/false,
        "explanation": "Your explanation here"
    }
```
\end{lstlisting}
    \caption{\ours\ system prompt to categorize the error case of false negatives.}
    \label{fig:fig_system_prompt_cat_fn}
\end{figure}

\begin{figure}[!h]
    \centering
\begin{lstlisting}
**Question**
{question}

**Prediction Query**
{pred_query}

**Prediction Result**
{pred_result}

**Ground Truth Query**
{gt_query}

**Ground Truth Result**
{gt_result}
\end{lstlisting}
    \caption{\ours\ user prompt. Execution results are excluded for generated queries that do not produce the same results as ground truth.}
    \label{fig:fig_user_prompt}
\end{figure}

% \begin{figure}[!h]
%     \centering
%     \lstinputlisting{prompt/prometheus_ambiguous}
%     \caption{Ambiguous criteria and rubrics utilized in \prometheustwo.}
%     \label{fig:fig_prometheus_criteria_ambiguous}
% \end{figure}

\begin{figure}[!h]
    \centering
\begin{lstlisting}
Criteria: How accurately does the Prediction Query answer the Question and adhere to SQLite3 compatibility?
Score 1: The Prediction Query is entirely incorrect, failing to answer the Question logically or containing critical errors that render it unusable. It may also be incompatible with SQLite3.
Score 2: The Prediction Query has significant flaws, such as missing essential tables/columns, incorrect filtering, or JOIN redundancy. It may partially answer the Question but has major issues. SQLite3 compatibility might be questionable.
Score 3: The Prediction Query somewhat answers the Question but has noticeable issues. It may fail to handle null values properly, misuse aggregation functions, or not consider multiple row scenarios. It is mostly SQLite3 compatible but may have minor issues.
Score 4: The Prediction Query largely answers the Question correctly with only minor flaws. It handles most scenarios well, including null values and multiple rows. It is fully SQLite3 compatible. Small oversights like missing DISTINCT in SELECT are acceptable.
Score 5: The Prediction Query perfectly answers the Question, considering all aspects including proper handling of null values, multiple rows, and edge cases. It is fully SQLite3 compatible and optimized. Minor differences like missing DISTINCT in SELECT are allowed if they don't affect the result.
\end{lstlisting}
    \caption{Criteria and rubrics utilized in \prometheustwo~for the equivalent set.}
    \label{fig:fig_prometheus_criteria_custom_eq}
\end{figure}
\begin{figure}[!h]
    \centering
\begin{lstlisting}
Criteria: How accurately does the Prediction Query answer the Question compared to the Ground Truth Query, considering logical correctness and result similarity?
Score 1: The Prediction Query completely fails to answer the Question logically. It contains significant errors, produces drastically different results from the Ground Truth Query, or fails to handle critical aspects like null values or multiple rows.
Score 2: The Prediction Query partially answers the Question but has major flaws. It may have incorrect filtering, missing conditions, or JOIN redundancy that significantly affect the results. The output differs notably from the Ground Truth Query in ways that impact the answer's accuracy.
Score 3: The Prediction Query generally answers the Question, but with some inconsistencies. It may have minor issues in handling null values or multiple rows. The results differ from the Ground Truth Query in ways that slightly affect the answer's completeness or accuracy.
Score 4: The Prediction Query correctly answers the Question with only minor differences from the Ground Truth Query. It handles null values and multiple rows appropriately. Any differences in output structure or value representation do not significantly impact the answer's accuracy.
Score 5: The Prediction Query perfectly answers the Question, either matching the Ground Truth Query or providing an equally valid alternative. It handles all aspects correctly, including null values and multiple rows. Any differences in output structure, column naming, or value representation are logically justified and do not affect the answer's accuracy
\end{lstlisting}
    \caption{Criteria and rubrics utilized in \prometheustwo~for the not equivalent set.}
    \label{fig:fig_prometheus_criteria_custom_neq}
\end{figure}
\twocolumn

\end{document}